\documentclass[12pt]{article}

\usepackage{amsmath, amsfonts, amssymb, dsfont, xcolor, comment}
\usepackage{soul}
\usepackage{natbib}
\usepackage{epsfig}
\usepackage{setspace}
\usepackage{graphicx, subcaption}
\usepackage{mdframed}
\usepackage{float}
\usepackage[super]{nth}
\usepackage[toc,page]{appendix}
\PassOptionsToPackage{hyphens}{url}\usepackage{hyperref}

\usepackage{subcaption}  
\usepackage[margin=1.25in]{geometry}
\usepackage[normalem]{ulem}
\usepackage[mathscr]{eucal}
\usepackage{booktabs}

\usepackage{longtable}
\usepackage{array}
\usepackage{ragged2e}
\usepackage{adjustbox} 

\newcommand{\Xomit}[1]{}
\newcommand{\superscript}[1]{\ifmmode^{#1}\else\(^{#1}\)\fi}

\long\def\/*#1*/{}

\begin{document}

\title{Divergent LLM Adoption and Heterogeneous Convergence Paths in Research Writing\thanks{We thank Yian Yin for the numerous discussions and detailed comments since the very inception of the project.
We also thanks for the excellent research assistance from Anqi Li,  Meixuan Wang, and Shuhuai Zhang.}}

\date{August 2024}
\author{Lin William Cong\footnote{Cornell University SC Johnson College of Business(Johnson) and NBER, will.cong@cornell.edu} 
    \and Wu Zhu\footnote{Tsinghua University, School of Economics and Management, zhuwu@sem.tsinghua.edu.cn}
}

\maketitle 

\begin{abstract}
Large Language Models (LLMs), such as ChatGPT, are reshaping content creation and academic writing. This study investigates the impact of AI-assisted generative revisions on research manuscripts, focusing on heterogeneous adoption patterns and their influence on writing convergence. Leveraging a dataset of over 627,000 academic papers from arXiv, we develop a novel classification framework by fine-tuning prompt- and discipline-specific large language models to detect the style of ChatGPT-revised texts. Our findings reveal substantial disparities in LLM adoption across academic disciplines, gender, native language status, and career stage, alongside a rapid evolution in scholarly writing styles. Moreover, LLM usage enhances clarity, conciseness, and adherence to formal writing conventions, with improvements varying by revision type. Finally, a difference-in-differences analysis shows that while LLMs drive convergence in academic writing, early adopters, male researchers, non-native speakers, and junior scholars exhibit the most pronounced stylistic shifts, aligning their writing more closely with that of established researchers. 

\end{abstract}

 \smallskip


\vspace{1cm}

\onehalfspacing

\newpage

\section{Introduction}
Since their inception in late 2022, Large Language Models (LLMs) such as ChatGPT have revolutionized many aspects of scientific research, with profound implications for academic writing \cite{pividori2023publishing}, scientific communication \cite{alvarez2024science,biyela2024generative}, discovery \cite{blanco2023role}, and productivity \cite{dwivedi2023opinion,khlaif2023potential}.
Recent studies have discovered a wide range of aspects that LLM can improve or accelerate \cite{thirunavukarasu2023large,yan2024promises}. Yet alongside these exciting potentials, cautionary tales  have emerged about how such tools inadvertently contaminate scientific production and communication like the LLM hallucination \cite{bang2023multitask},misinformation \cite{bontridder2021role,else2023chatgpt,blanco2023role,alvarez2024science,garry2024large}, and ethical issues \cite{dergaa2023human,thorp2023chatgpt}. Together, these developments raise a series of open questions regarding the fundamental challenges and opportunities that LLMs present for science and innovation. 
Despite growing discussions on the impact of large language models (LLMs) on science, a fundamental question remains underexplored: \textit{How have scientists utilized LLMs to write scientific papers, if at all?} While several researchers have developed models to detect the presence of GPT or generative AI in academic writing, their focus primarily lies in identifying whether text was machine-generated \cite{gehrmann2019gltr,solaiman2019release,chakraborty2023possibilities,mitchell2023detectgpt,gao2023comparing,liang2024monitoring}. However, limited attention has been given to understanding \textit{how} GPT is used and its influence on writing style. Moreover, there is a paucity of analysis regarding the heterogeneous adoption of these tools across diverse researcher groups and their evolving impact on scientific communication.

To detect whether a text has been revised using GPT in a specific style, we developed a robust framework for systematically generating AI-revised text through targeted prompts. We compiled a comprehensive dataset of 627,384 academic papers posted on arXiv between January 2021 and December 2023. This dataset includes detailed metadata for each article, such as abstracts, fields of study, authorship information, and revision histories. Spanning both pre- and post-GPT adoption periods, it enables us to capture the evolving influence of GPT over time.

Using zero-shot prompting techniques \cite{brown2020language,wei2022chain}, we instructed GPT-3.5 to revise human-written abstracts along key dimensions, including clarity, formality, objectivity, readability, and grammatical correctness, while preserving the original meaning of the text. These revisions formed the basis for training and fine-tuning 48 field- and prompt-specific models to distinguish between human and GPT-revised abstracts. To ensure unbiased model performance, we used articles written before November 2022, prior to GPT’s widespread release, as ground truth for human-written content. This approach safeguarded the integrity of our training data and ensured that our models accurately identified AI-revised text without contamination from GPT-generated revisions.

Our models are trained using a large dataset of human-written abstracts paired with their corresponding GPT-revised versions, achieving state-of-the-art classification accuracy. With out-of-sample precision, recall, and F1 scores consistently exceeding 96\%, these models offer a highly reliable framework for identifying GPT-revised texts. This approach not only enables the estimation of GPT usage prevalence across academic disciplines but also facilitates an in-depth understanding of the specific revisions applied to scientific writing. Through this framework, we contribute a robust tool for analyzing the evolving impact of GPT on academic communication.

A detailed examination of GPT adoption in academic writing revealed significant heterogeneity across disciplines. By the end of 2023, approximately 22\% of abstracts in Computer Science were revised using GPT, with Electrical Engineering and Systems Science (EE\&SS) following closely at 21\%. Adoption rates in Finance and Statistics were 18\% and 15\%, respectively, while Biology and Economics each showed adoption rates of around 11\%. Physics exhibited a lower adoption rate of 10\%, and Mathematics demonstrated the least uptake, with only 3.5\% of abstracts revised using GPT. These disparities suggest that disciplines more engaged in computational methods are more inclined to adopt AI-driven tools for refining research papers.

In addition to disciplinary differences, we observed substantial heterogeneity in GPT adoption based on access, seniority, and researcher demographics. Junior researchers and non-native English speakers were significantly more likely to use GPT for revisions compared to their senior and native-speaking counterparts. Variations in access to GPT across regions and institutions further contributed to these differing adoption patterns. 

Meanwhile, the evolution of GPT usage over time highlights a shift in its application. Initially, researchers primarily used GPT to enhance clarity and conciseness in abstracts. For instance, in the first quarter of 2023, 45\% of articles revised by GPT focused on improving clarity and conciseness, and 22\% aimed at correcting grammar and syntax. By the fourth quarter, these figures declined to 33\% and 9.7\%, respectively, as more researchers began leveraging GPT for comprehensive revisions addressing multiple aspects of writing. During this same period, the share of comprehensive revisions increased from 19.5\% to 27.7\%, reflecting a growing reliance on GPT for advanced writing enhancements.

The rapid adoption of Large Language Models (LLMs) such as GPT in scientific writing has opened new possibilities for enhancing writing quality. However, it raises a critical question: how does GPT-assisted writing influence academic writing style and quality? While prior research has examined various aspects of scientific writing and its relationships with knowledge dissemination, publication, and citation \cite{chiswick2004writing,weinberger2015ten,dejesus2019generic,forero2020brief,kousha2024factors}, our study extends these foundations by directly comparing human-written and GPT-revised abstracts across key textual dimensions and evaluating their real-world impact. Specifically, we aim to quantify the influence of GPT on writing by analyzing texts using a set of basic writing principles known to improve publication and citation outcomes \cite{weinberger2015ten,kousha2024factors}.

To estimate the effect of GPT revisions, we conducted a controlled experiment comparing original arXiv abstracts with their GPT-revised counterparts. The comparison focused on ten key writing principles, including brevity, clarity, the use of present tense, avoidance of adjectives and hedge words, and signaling importance and confidence. We employed a linear regression model with article-level fixed effects, controlling for characteristics such as research field, topic, and content. This approach enabled us to measure the direct impact of GPT revisions within each article while mitigating potential selection bias arising from researchers' decisions to use GPT.

Our analysis reveals three key insights into the impact of GPT on academic writing. First, GPT revisions significantly enhance conciseness and clarity, often reducing word count by over 25\%. Second, GPT shifts writing styles toward greater formality, favoring present tense and minimizing the use of adjectives, adverbs, and hedge words, resulting in a more neutral tone. Third, GPT revisions align closely with the structured writing preferences of senior researchers, suggesting that GPT may standardize writing styles across experience levels. While these findings underscore GPT's potential to improve writing quality, they also raise concerns about the homogenization of academic expression.

To further investigate the impact of GPT-assisted writing on style convergence among different researcher groups, we examined how GPT adoption influences writing patterns. Specifically, we explored whether widespread GPT usage leads to a more uniform style across researchers of varying backgrounds. As part of this analysis, we conducted a thought experiment comparing original arXiv abstracts with their GPT-revised counterparts, employing cosine similarity metrics to quantify the degree of convergence. This approach allowed us to measure changes in writing style across groups based on seniority, gender, and nativeness.

Using a difference-in-difference analysis, our results reveal that GPT-assisted revisions significantly drive convergence in writing styles. These revisions produce texts that are more concise, clear, and professionally polished, simplifying complex language and enhancing readability. This convergence is particularly pronounced between junior and senior researchers, with junior scholars increasingly producing work that mirrors the polished style of more experienced authors. As LLMs become more widely adopted, we observe a trend toward standardized writing practices, potentially democratizing academic publishing by improving the writing quality of less experienced researchers. This finding contrasts with concerns that GPT usage might exacerbate socioeconomic inequalities due to uneven benefit distribution \cite{liang2024can,capraro2024impact,yu2024whose,wilmers2024generative}. However, this shift also raises questions about the potential loss of diversity in scientific expression.

Convergence is especially evident among researchers who actively adopt GPT for revisions. Both junior and senior authors using GPT exhibit a marked shift toward GPT-style writing after the public release of ChatGPT, whereas those who do not adopt GPT maintain more stable writing patterns. Furthermore, the degree of convergence varies across gender and nativeness. Male and non-native researchers who adopt GPT show higher levels of style convergence, while female authors demonstrate less pronounced changes. These findings suggest that while GPT improves writing quality, it may also contribute to the homogenization of writing styles, particularly among researchers who rely heavily on the technology.

\paragraph{Literature.} 
Our paper makes a significant contribution to the growing literature on the role of generative AI in scientific research. For instance, \cite{pividori2023publishing} examines how LLMs improve clarity and quality through case studies in medicine. Similarly, \cite{alvarez2024science} explores the potential and challenges of generative AI in science communication, emphasizing its role in translating and summarizing complex scientific information. \cite{else2023chatgpt} highlights the difficulty researchers face in distinguishing ChatGPT-generated abstracts from human-written ones, while \cite{le2023chatgpt} evaluates GPT adoption through surveys targeting postdoctoral researchers. Additionally, \cite{pividori2023publishing} underscores the importance of prompt engineering in scientific writing, and \cite{liang2024can} demonstrates GPT-4’s utility in providing scientific feedback by analyzing overlaps between human- and GPT-generated feedback on articles from \textit{Nature} and the ICLR conference.

Unlike prior studies that rely heavily on case studies or surveys, our paper is the first to systematically examine the use of generative AI in academic revisions at scale and its heterogeneous impacts across researcher groups. Leveraging a comprehensive dataset of all articles posted on arXiv, our analysis generalizes findings beyond the limitations of small-scale or domain-specific studies. By identifying how GPT adoption affects writing style, quality, and convergence across disciplines and researcher demographics, our study provides a broader and more nuanced understanding of generative AI’s transformative role in academic publishing.

Our paper also contributes to the growing body of literature on detecting machine-generated text. The proliferation of generative AI has raised widespread concerns about its potential to contaminate human-authored content, as distinguishing LLM-generated text from human-written content becomes increasingly challenging \cite{clark2021all,gao2023comparing,else2023chatgpt,liang2024monitoring}. Existing approaches to LLM detection primarily focus on identifying AI-generated text at the document level using zero-shot or self-detection methods. These approaches leverage the observation that LLM-generated texts exhibit distinctive probabilistic or geometric patterns within the generating model \cite{bakhtin2019real,zellers2019defending,uchendu2020authorship}. For instance, \cite{mitchell2023detectgpt} developed DetectGPT, a zero-shot detector that utilizes curvature properties of log probabilities to identify machine-generated text. 

Another prominent detection approach involves fine-tuning pretrained models on datasets containing both human- and AI-generated text to distinguish between the two, eliminating the need for access to the original generating model. Our paper adopts this methodology but focuses on discipline- and prompt-specific models to identify specific writing styles. For example, \cite{chen2023gpt} developed GPT-Sentinel, a model fine-tuned on RoBERTa \cite{liu2019roberta} and T5 \cite{raffel2020exploring}, using a dataset of human-written and GPT-rephrased content to detect ChatGPT-generated text. Similarly, \cite{yu2023gpt} introduced LLM-Pat, which employs an intermediary LLM to reconstruct and compare texts for machine-generated content identification. Additionally, \cite{liang2024monitoring} conceptualized text as a composition of human- and machine-generated components and developed a model to estimate the proportion of machine-generated text. Other researchers have proposed text watermarking as a detection method, embedding algorithmically detectable signals within AI-generated text \cite{khlaif2023potential,zhao2023provable,wu2023dipmark,fernandez2023three}.

Despite these advancements, existing studies predominantly address the binary question of whether text is human- or machine-generated, without delving into \textit{how} or \textit{in what style} the text was generated by the machine. Our paper advances this literature by developing a set of discipline- and prompt-specific LLMs to detect how researchers use generative AI to rewrite their articles. Furthermore, we examine the impact of these revisions on writing styles across different disciplines, providing novel insights into the intersection of AI and academic communication.

\section{Data and LLMs}

To provide quantitative insights into the detection of LLM-generated scientific text, we developed a systematic pipeline to identify scientific abstracts likely produced by large language models. We began by constructing a ground truth dataset of human-written scientific text, comprising 343,461 articles posted on arXiv during the pre-ChatGPT era (January 1, 2021, to October 31, 2022). Using zero-shot prompts, we instructed GPT-3.5-Turbo to produce AI-revised versions of these texts. Specifically, for each human-written abstract, we directed GPT to revise and enhance the text according to various academic writing instructions while preserving its core meaning. 

Building on these corpora, we trained and fine-tuned 48 field- and prompt-specific large language models capable of classifying scientific abstracts as either GPT-generated or human-written. Testing this framework on over 620,000 preprints posted on arXiv between January 1, 2021, and December 31, 2023, our models achieved a high level of accuracy in detecting LLM-assisted abstracts, demonstrating the robustness and scalability of this approach.

While similar approaches have been employed in recent studies of LLM impact on science \cite{chen2023gpt,liang2024can}, our implementation distinguishes from existing tools by explicitly examining different potential uses of LLM and their potential and heterogeneous impact on a variety of researchers. In particular, during the revision stage, we instruct the GPT-3.5 to revise an abstract along different dimensions: 1) Enhance Clarity and Conciseness, 2) Increase Formality and Professionalism, 3) Improve Objectivity, 4) Enhance the Readability and Understandability, 5) Correct Grammar and Syntax Errors, and 6) Make a Comprehensive Improvement encompassing all the above five aspects. Appendix A contains the specific prompts used to generate these revisions.

Table \ref{tab3:differnce-bt-ai-human} presents an example based on a human-written abstract, highlighting substantial differences across GPT-rewritten versions. For example, Revision 1 eliminated redundant phrases, simplified complex sentences, and focused on the main points to make the abstract more direct and easier to read, changed "combines" to "merges", and "will be" to "is" etc, and cut the number of words from 194 to 173 and unchanged in the number of sentences. Revision 2 introduced more formal language, included precise scientific terminology, and structured sentences to reflect scholarly writing. For example, it changed the sentence "However, it is empirically known that Adam often generalizes worse than..." to a more formal version "Notably, Adam is empirically understood to generalize less effectively than...", cut the number of words to 172 and unchanged the number of sentences.  Revision 3 modified subjective statements to be more fact-based, used passive constructions where appropriate, and emphasized results and methodologies rather than perceptions or beliefs.
It revised the sentence "The purpose of this paper is to unveil the mystery of this behavior..." to "This paper aims to clarify this disparity through a diffusion theoretical approach...". "Aim to clarify" is more objective and less dramatic than "unveil the mystery," aligning better with scientific writing standards. Besides, it also cut the number of words significantly to 163. Revision 4 simplified complex vocabulary, broke down intricate descriptions, and clarified scientific concepts without assuming extensive prior knowledge. For example, "...uses parameter-wise adaptive inertia to accelerate the training and provably favors flat minima as well as SGD." was revised to "...uses adaptive inertia tailored to each parameter to speed up training and select flat minima as effectively as SGD.". "Tailored to each parameter" and "speed up training" are simpler and easier to understand than "parameter-wise adaptive inertia" and "accelerate the training.". Besides, it also cut significantly the number of words to 162. Revision 5 scanned the text for grammar mistakes, punctuation errors, and syntactical awkwardness, and corrected them to ensure the language flows smoothly. For instance, the original phrase "...and almost does not affect flat minima selection." was revised to "...and barely influences minima selection.", and the text was cut to 172 words. The final revision integrated the enhancements made in clarity, formality, objectivity, readability, and grammar into one cohesive and refined piece. For example, in terms of formality, it changed the "would be the most popular" to "is a premier" indicating a more assertive and professional tone. In terms of clarity and objectivity, "The purpose of this paper is to unveil the mystery" was transformed to "This study delves into this issue", which is clearer and more objective. In terms of conciseness and readability, several phrases were streamlined for brevity and ease of understanding, such as "which uses parameter-wise adaptive inertia" simplified to "This approach leverages parameter-specific adaptive inertia."

\begin{figure}[H]
\begin{center}
\caption{Vector Representation of Various Versions}  
    \vspace{-2pt}
    \label{figure2_embedding_difference}
    \includegraphics[width=1\textwidth, height=0.75\textheight]{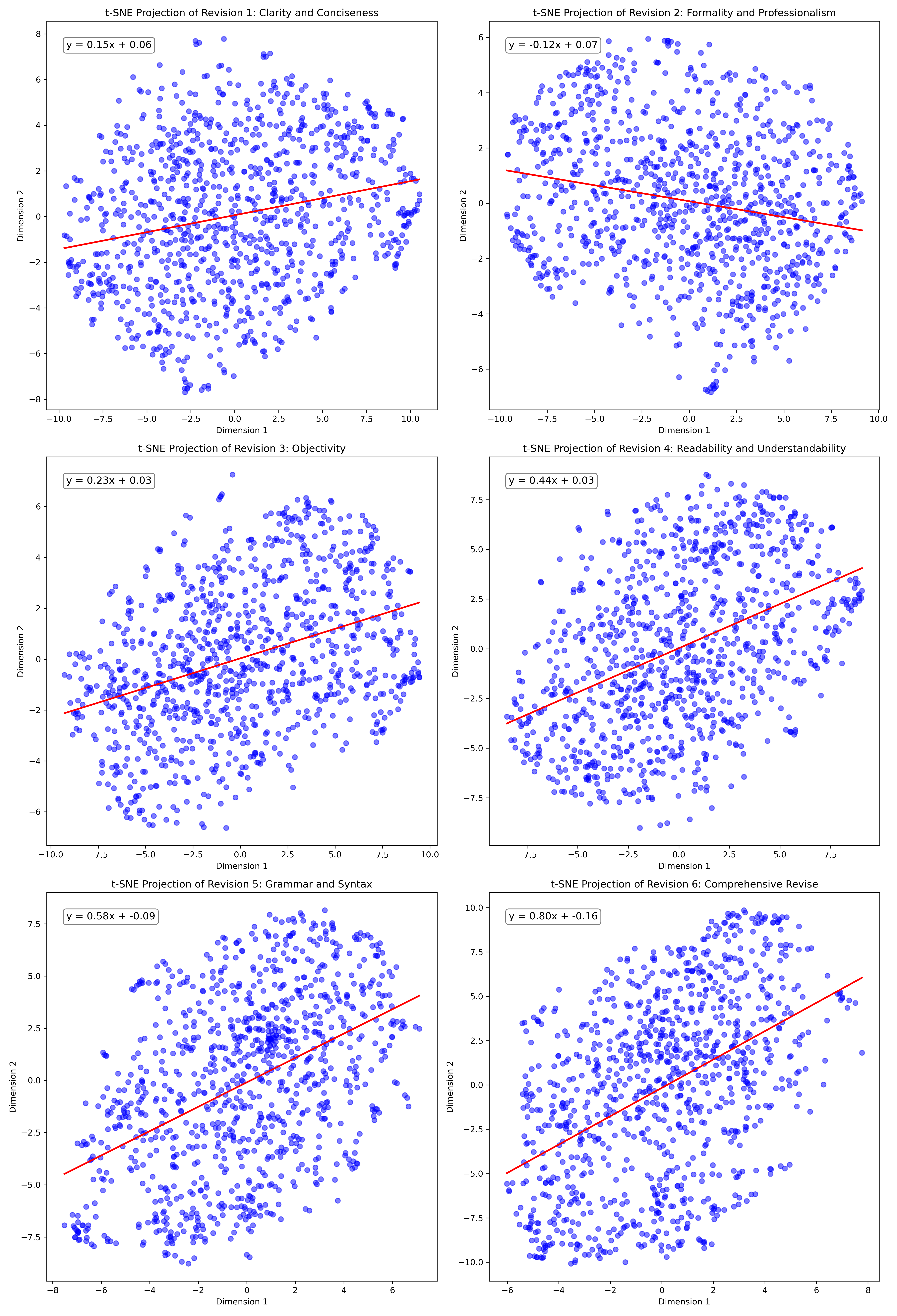}
    \vspace{-8pt}
\end{center}   
{\footnotesize 
Note: This figure illustrates the semantic differences between versions. We randomly sampled 1,000 articles and generated embeddings using the BERT model. We calculated the difference between each article and its revised version for their vector embeddings. The t-SNE algorithm was then applied to project the 768-dimensional vectors into a two-dimensional subspace, preserving the two most important components. The figure demonstrates that the revised versions differ significantly from the original abstracts, highlighting noticeable variations across revisions.
}   
\end{figure}

To investigate whether there are systematic differences among revisions, we utilize paragraph embeddings to represent them. Figure \ref{figure2_embedding_difference} illustrates the differences in vector representations for various types of revisions. Specifically, we select a random sample of 1,000 abstracts, transforming each into a 768-dimensional vector using BERT, a widely adopted model for capturing semantic relationships between sentences and paragraphs \cite{devlin2018bert,reimers2019sentence}. 

For each abstract, we calculate the vector difference between its original content and the revised version, representing the semantic shift induced by the revision. These differences are then visualized in a two-dimensional subspace to better understand their patterns. To achieve this, we apply the t-SNE algorithm (t-distributed Stochastic Neighbor Embedding) \cite{van2008visualizing}, a non-linear dimensionality reduction technique designed to preserve local structure in high-dimensional data while projecting it into a lower-dimensional space. By focusing on the two most important components, t-SNE enables the detection of patterns and clusters, making it an effective tool for visualizing these revision-induced changes.

Several noteworthy observations emerge from the analysis. First, the distribution of the projections does not collapse into $(0,0)$, indicating that the revised versions are semantically distinct from the original content. Second, the joint distribution of the two key coordinates differs across revisions. For instance, the first revision (relative to the original content) is confined to the region $[-10,10] \times [-8,8]$, whereas the sixth revision is concentrated within $[-6,8] \times [-10,10]$. Finally, the relationship between the first and second coordinates varies, as reflected in the differing slopes of the fitted lines for each revision. Collectively, these distinctions highlight the semantic shifts induced by revisions, providing a basis for developing large language models (LLMs) capable of distinguishing between GPT-revised and human-revised texts, as well as between different types of revisions.

The semantic differences captured by the vector representations further enable us to design field- and prompt-specific LLMs that differentiate between original abstracts and their GPT-revised versions. These models, as discussed in detail in the next subsection, leverage these distinct embeddings to achieve precise classification.

\subsection{Mothods}
This section describes in detail the text dataset, the process pipeline for obtaining human-written texts which GPT then revises following different prompts, and the LLMs developed to differentiate various revisions. The metadata includes article ID, abstract, field code, revision dates, authors, and their affiliations and countries. Based on the field code, we categorize the articles into eight fields: Mathematics, Physics, Computer Science, Electrical Engineering and Systems Science, Biology, Economics, and Finance. Table \ref{tab1_number_of_articles} presents the number of unique articles published on the arXiv, between 2021 and 2023 by academic disciplines. 


We re-trained a set of field- and prompt-specific large language models (LLMs) to identify whether abstracts were written by GPT or human researchers. Our identification strategy leverages the release timeline of ChatGPT-3.5. Specifically, ChatGPT was first released by OpenAI in November 2022 and became publicly accessible for academic text revisions thereafter. Consequently, we are confident that articles last updated before October 31, 2022, could not have been written or revised by ChatGPT-3.5 (online version). For these articles, we label the original abstract from arXiv as human-written, while the revised versions are labeled as GPT-written.

To train the model, we use articles with their latest updates before October 1, 2021, as training data. K-fold cross-validation (with \( K = 5 \)) is applied to optimize hyperparameters and implement early stopping, saving the best model when no further improvement in cross-validation accuracy is observed. Articles updated between October 1, 2021, and November 30, 2021, are used as test data for out-of-sample evaluation. We restrict our analysis to data before November 30, 2021, to ensure no abstracts were revised using GPT-3, as GPT-3 was publicly released only after this period.


To differentiate between human-written and GPT-revised texts, we fine-tuned a well-pretrained base model, \textit{all-mpnet-base-v2}, with a classifier sub-module for the final downstream classification task. This pretrained model is a sentence-transformer developed by Microsoft that achieves state-of-the-art performance in sentence representation. It maps sentences and paragraphs to a 768-dimensional dense vector space, suitable for tasks such as clustering and semantic search. We fine-tuned the model to accommodate our classification problem. 

The customized classifier sub-module includes a varying number of layers (one to three), a varying number of neurons per layer, and dropout rates, all of which are optimized using five-fold cross-validation. Our classification pipeline involves two distinct phases. First, we fine-tuned discipline- and prompt-specific models to differentiate human-written texts from GPT-revised texts following specific prompts, resulting in 48 models for binary classification (0-1). Second, we fine-tuned discipline-specific models to differentiate human-written texts from revisions based on different prompts, producing 8 multiclass models (7 classes: 0 for human-written and 1–6 for different revision types).

Table \ref{tab2_out of sample performance evaluation}  summarizes the model performance for binary classification on training and testing samples. We report four key metrics: precision, recall, accuracy, and F1 score. \textbf{Precision} measures the percentage of paragraphs identified as GPT-revised that were actually revised by GPT. \textbf{Recall} measures the percentage of GPT-revised texts correctly identified by the model. \textbf{Accuracy} represents the overall percentage of correctly classified texts, and the \textbf{F1 score} is the harmonic mean of precision and recall. Our method achieves state-of-the-art classification accuracy, ranging from 96.9\% for Revision 5 to nearly 99.6\% for Revision 3. Figure \ref{figure_classification_accuracy} in Appendix A demonstrates that our model effectively separates unrevised and revised texts in terms of their distribution.

To quantify the accuracy of our model in classifying each article, we estimate two key probabilities: \( P(\hat{y}_i = 1 \mid y_i = 0) \), the probability of incorrectly classifying a human-written article as GPT-revised, and \( P(\hat{y}_i = 1 \mid y_i = 1) \), the probability of correctly identifying a GPT-revised article. Figure \ref{figure_classification_accuracy} in Appendix A illustrates the distribution of these probabilities for each text. The distributions are well-separated, concentrated, and far from 0.5, indicating that our trained models can reliably and accurately classify articles.

Beyond accuracy, training on GPT-revised texts across different revision dimensions enables our classifiers to go beyond binary detection. Specifically, our models can estimate the probability of LLM use in scientific papers while also inferring the likely objectives and prompts guiding such usage. This functionality provides a nuanced understanding of how researchers utilize LLMs in revising scientific content.

Table \ref{tab4_confusion_matrix_multiclasses} presents the confusion matrix for the multi-class classification task. Specifically, the entry \((i,j)\) in the \(i\)th row and \(j\)th column represents the percentage of texts with a ground truth revision of \(i\) being identified as revision \(j\), where revision 0 corresponds to the original abstract. Several noteworthy observations emerge from this table.

First, the model is highly effective at distinguishing unrevised abstracts from GPT-revised versions. For instance, 92.25\% of the original (unrevised) texts are correctly classified as such, and nearly all revised texts are accurately recognized as revisions, as indicated by the near-zero percentages in rows 2–6 of the first column.

Second, the model performs well in differentiating revisions 1, 2, 3, and 6 from other types of revisions but exhibits challenges in reliably distinguishing revisions 4 and 5. Specifically, only 23.76\% of revision 4 texts and 18.46\% of revision 5 texts are correctly classified. This is likely due to the nature of these revisions. Revision 4 focuses on enhancing readability and understandability, while revision 5 addresses grammar and syntax corrections. These types of changes often overlap with other revision categories, especially as our prompts guide GPT to refine the text for submission to top conferences or journals in addition to specific revision objectives. As a result, the nuanced improvements associated with revisions 4 and 5 may share similarities with other types of textual modifications, leading to lower classification accuracy for these categories.

\section{GPT in Academic Writing}\label{sec:gpt_adoption}

\subsection{Human vs. AI Writing}
This section documents the growing trend of GPT being used for revising or writing abstracts across academic disciplines. In addition, it highlights significant heterogeneity in GPT adoption across various factors, including discipline, native language, country, gender, and research experience.

\subsection{Adoption of GPT in Writing}
\begin{figure}[H]
\begin{center}
\caption{Percentage of Texts Identified as being Revised by ChatGPT}  
    \vspace{-2pt}
    \label{figure3_percentage_of_gpt}
    \includegraphics[width=1\textwidth, height=0.75\textheight]{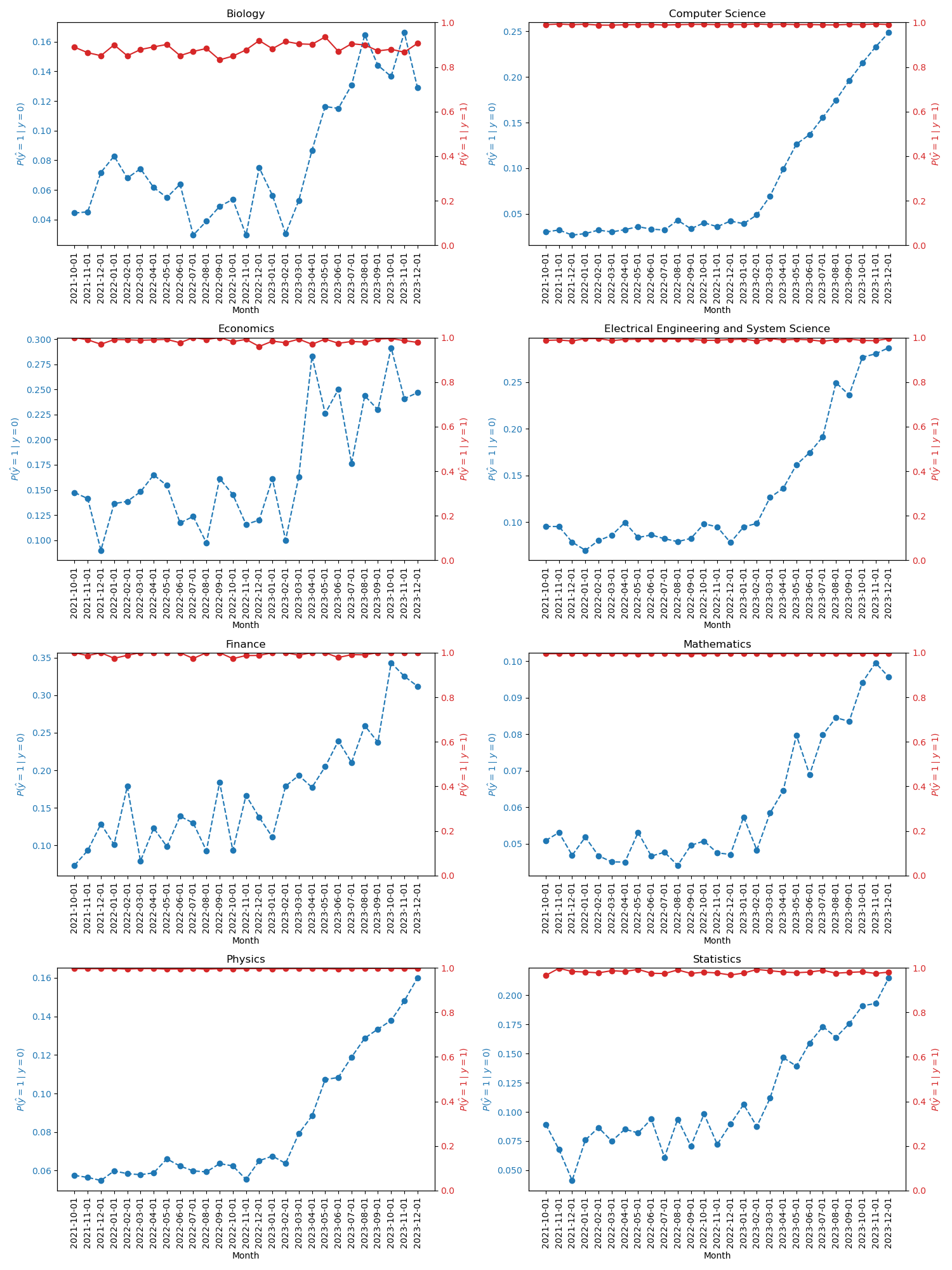}
    \vspace{-8pt}
\end{center}   
{\footnotesize 
Note: This figure illustrates the monthly percentage of articles identified as having been revised by GPT-3.5 across eight disciplines. The analysis focuses on the binary (0-1) classification between the original abstracts and their sixth revision, with similar trends observed for other revisions. The blue dashed line represents the monthly percentage of original abstracts identified as being revised by GPT-3.5 toward version 6. In contrast, the red solid line shows the monthly percentage of version 6 abstracts identified as being revised by GPT-3.5, aligned with the direction of version 6.
}   
\end{figure}

The distinct differences in writing styles between GPT and human researchers enable the identification of human-written and GPT-revised texts across various dimensions. This subsection highlights the widespread adoption of GPT in academic writing, particularly across different fields and research communities. Using our fine-tuned model, we can detect whether GPT was used to revise an abstract and specify the direction of the revision. Figure \ref{figure3_percentage_of_gpt} shows the percentage of abstracts revised by GPT across eight disciplines, revealing significant variations in GPT adoption rates.

Several key observations are noteworthy. First, prior to November 30, 2022, the percentage of AI-revised articles remained relatively stable. This stability is primarily attributed to misclassification by the model, a common issue in the systematic detection of LLM-generated text \cite{chen2023gpt,liang2024monitoring} \footnote{The misclassification error is inversely related to the sample size. For instance, accuracy is higher in disciplines like Computer Science, which have larger training datasets, while fields such as Economics and Finance, with smaller datasets, exhibit higher error rates.}.To address this issue, we adjusted the raw values by subtracting the average adoption rate observed before November 2022. This normalization ensures that the metric has an average value of zero prior to the release of ChatGPT, enabling a clearer interpretation of adoption trends post-release.

Following the public release of ChatGPT, a rapid and substantial increase in AI-generated abstracts was observed across all academic disciplines. Within the first year, the proportion of AI-written abstracts surged dramatically, rising from 0.47\% in January 2023 to 13.8\% in December 2023—a 740-fold increase. This remarkable growth in adoption, however, exhibited significant variation across fields.

By the end of 2023, the share of AI-generated abstracts increased by 22\% in Computer Science, 21\% in Electrical Engineering and Systems Science (EE\&SS), 18\% in Finance, 15\% in Statistics, 11\% in both Economics and Biology, 10\% in Physics, and 3.5\% in Mathematics compared to the end of 2022. These field-specific differences highlight the uneven adoption of AI tools, with disciplines more closely tied to computational and quantitative methods demonstrating higher uptake.

\subsection{Diverse uses of GPT in scientific writing}

Given the diverse standards and practices of scientific writing across disciplines, we further investigate whether scholars from different fields utilize ChatGPT in similar or distinct ways. Leveraging the 56 classifiers trained earlier, we compare the distributions of prompt usage across various research communities. To analyze how abstracts are revised based on different prompts, we employ a multi-label classification model to identify the types of revisions applied. This approach enables us to capture the nuanced ways in which researchers adapt their writing to align with discipline-specific conventions and practices.

\begin{figure}[H]
\begin{center}
\caption{Percentage of Texts Identified being Revised by ChatGPT in various Dimensions.}  
    \vspace{-2pt}
    \label{figure4_percentage_of_gpt_in_six_types}
    \includegraphics[width=1\textwidth, height=0.75\textheight]{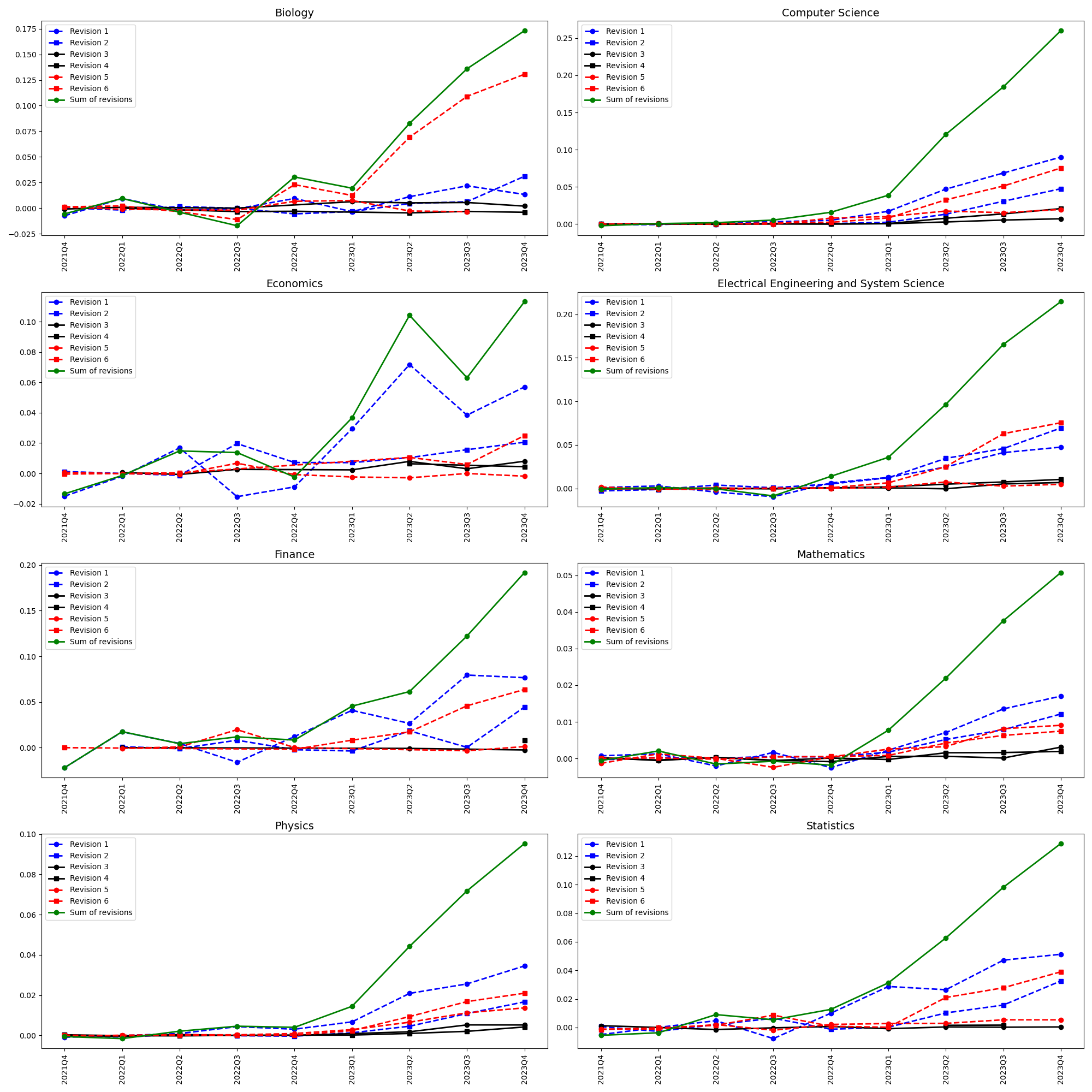}
    \vspace{-8pt}
\end{center}   
{\footnotesize 
Note: This figure displays the percentage of abstracts identified as having been written by ChatGPT, based on predictions from our trained multi-label large language model (LLM). Specifically, the model classifies the original abstract, where revision 0 represents the original version, revisions 1-5 correspond to versions revised along five distinct dimensions, and revision 6 represents a composite revision. The figure reports the percentage of articles revised by each of the six prompts, as well as the total percentage of articles revised (the sum of revisions 1 through 6). For clarity, we have subtracted the average percentage of revisions before November 2022, normalizing that period to zero.
}   
\end{figure}

Figure \ref{figure4_percentage_of_gpt_in_six_types} illustrates how different disciplines employ GPT to revise abstracts in distinct ways, highlighting significant variation in technology adoption across various aspects of writing. Between the first and fourth quarters of 2023, the percentage of texts revised according to prompt 1 increased from 1.1\% to 5.3\%, from 0.22\% to 3.03\% for prompt 2, from 0.06\% to 0.53\% for prompt 3, from 0.04\% to 1.03\% for prompt 4, from 0.54\% to 1.54\% for prompt 5, and from 0.46\% to 4.4\% for prompt 6, which corresponds to comprehensive revisions.

More importantly, GPT usage has evolved rapidly over time. Initially, researchers primarily used GPT to enhance clarity and conciseness in their abstracts. However, there has been a noticeable shift toward more comprehensive revisions addressing multiple dimensions simultaneously. For example, in the first quarter of 2023, approximately 45\% of articles revised using GPT focused on improving clarity and conciseness, while 22\% targeted grammar and syntax enhancements. By the fourth quarter of 2023, these figures had declined to 33\% and 9.7\%, respectively. Conversely, the proportion of articles employing GPT for comprehensive revisions increased significantly, rising from 19.5\% in the first quarter to 27.7\% in the fourth quarter. This trend underscores a growing reliance on GPT for advanced and multifaceted improvements in academic writing.

{\subsection{Individual characteristics and GPT adoption}}

We further investigate the relationship between adoption behavior and individual-level characteristics, including gender, race, ethnicity, career stage, and prior productivity. Across all examined dimensions, we observe systematic heterogeneity in both the popularity and timing of LLM adoption.

\begin{figure}[H]
\begin{center}
\caption{Native v.s. Non-Native in Using GPT to Revise Abstracts }  
    \vspace{-2pt}
    \label{figure5_native_vs_nonnative_gpt}
    \includegraphics[width=1.\textwidth, height=0.6\textheight]{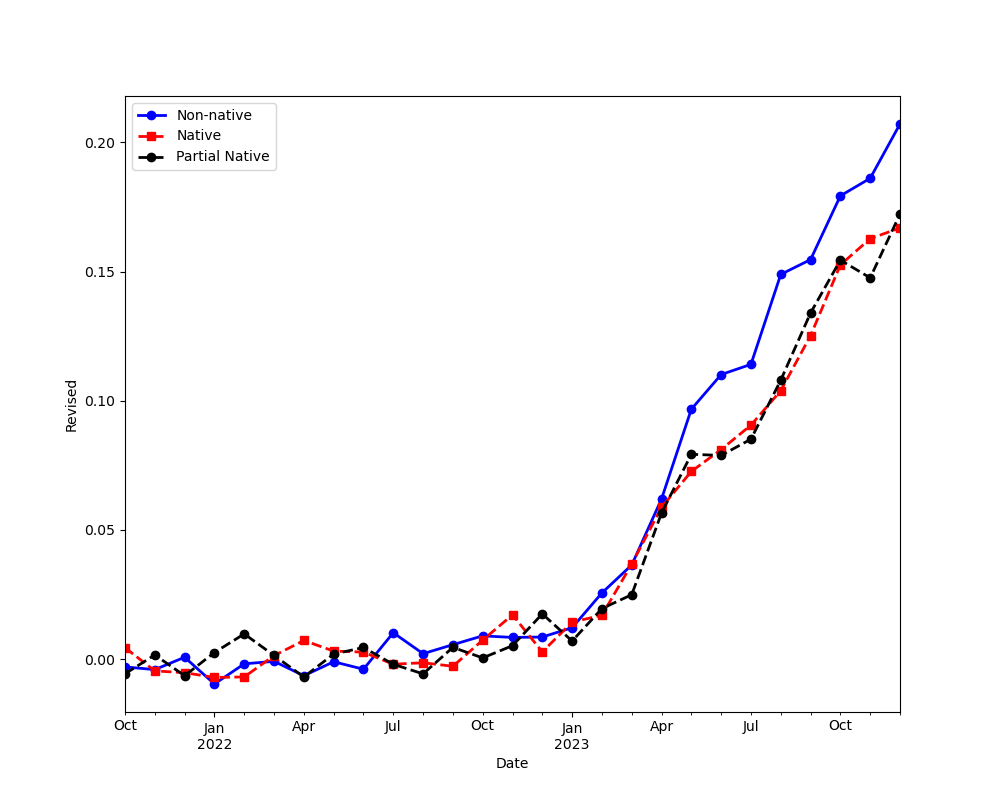}
    \vspace{-2pt}
\end{center}   
{\footnotesize 
Note: This figure illustrates the difference in GPT adoption for abstract revisions between native and non-native writers. An author is classified as native if their affiliation is in a country where English is the native language, including the United States, Australia, the United Kingdom, Canada, South Africa, and New Zealand. All other countries are classified as non-native. At the paper level, we calculate the percentage of native authors and categorize articles into three groups: native (all authors are native), non-native (all authors are non-native), and partial native (some authors are native). The x-axis represents the month, and the y-axis shows the percentage of articles revised by GPT. An abstract is considered revised by GPT if it is classified as 1, 2, 3, 4, 5, or 6 by our multi-label classification LLM.
}   
\end{figure}

Figure \ref{figure5_native_vs_nonnative_gpt} highlights the differences in GPT adoption between native and non-native speakers for abstract writing. Two key observations emerge. First, prior to the release of ChatGPT in November 2022, the percentage of articles identified as GPT-revised remained stable and near zero, after adjusting for baseline classification error, regardless of whether they were authored by native, non-native, or partially native speakers. Second, following ChatGPT's release, the percentage of GPT-revised abstracts increased sharply, particularly among non-native speakers, eventually surpassing the revision rate for native authors. This trend indicates that non-native speakers are more inclined to leverage GPT for abstract revisions post-release, suggesting a potential role for GPT in bridging linguistic barriers in academic writing.

\begin{figure}[H]
\begin{center}
\caption{Enthinicity Difference in Using GPT to Revise Articles }  
    \vspace{-2pt}
    \label{figure6_ethnicity_gpt}
    \includegraphics[width=1.\textwidth, height=0.6\textheight]{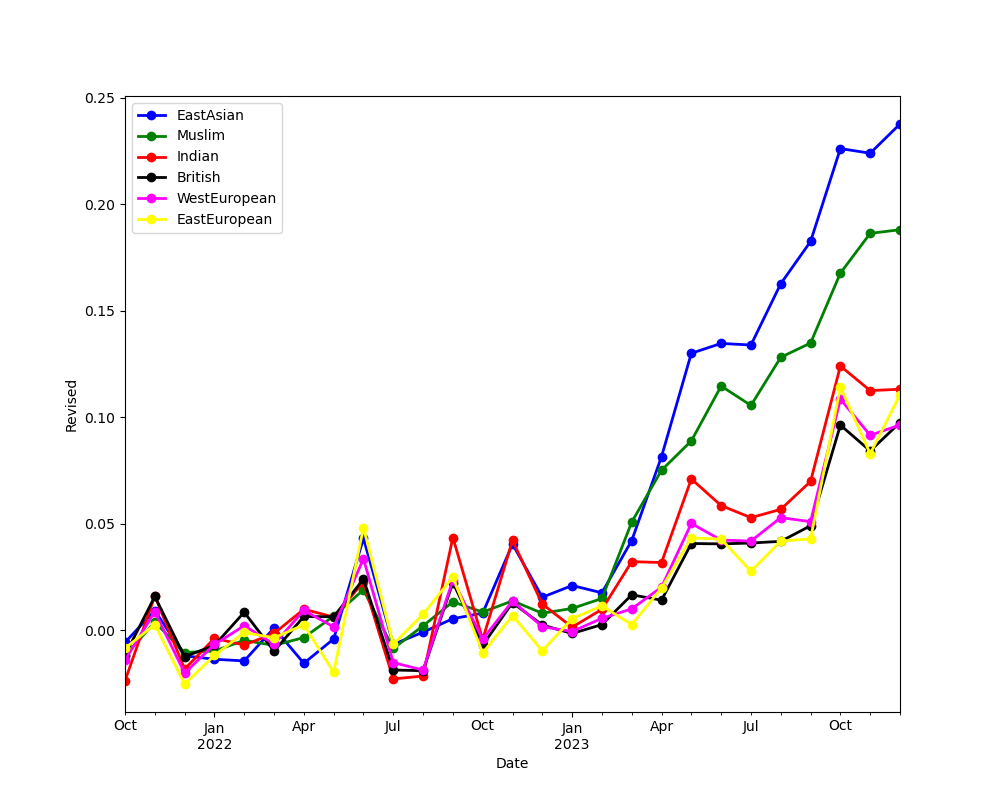}
    \vspace{-2pt}
\end{center}   
{\footnotesize 
Note: This figure illustrates the differences in GPT adoption for abstract revisions across different ethnicities. We use an online package \footnote{Specifically, the package is pred\_wiki\_name from the ethnicolr library.} to infer an author’s ethnicity based on their name. Authors are classified into several groups: East Asian, Muslim, Indian, British, West European, East European, and Other. We then calculate the percentage of articles written by GPT for each month and ethnicity. The x-axis represents the month, and the y-axis shows the monthly percentage of articles written by GPT.
}   
\end{figure}

Figure \ref{figure6_ethnicity_gpt} highlights the trend of abstracts identified as being revised by GPT across different ethnic groups. Consistent with previous findings, we observe stable and near-zero revisions prior to ChatGPT's release, followed by a significant surge in GPT usage for writing abstracts. This increase is particularly pronounced among academic researchers from East Asia (including China, Japan, and South Korea), with nearly 25\% of articles being revised by GPT. Muslim and Indian researchers follow, with approximately 18\% and 10\% of their abstracts being revised by GPT, respectively, echoing the patterns observed among non-native speakers (see Figure \ref{figure5_native_vs_nonnative_gpt}). In contrast, researchers from Europe and the United Kingdom exhibit the least inclination to revise their articles using GPT, indicating a slower adoption of the technology in these regions.

\begin{figure}[H]
\begin{center}
\caption{Gender Difference in ChatGPT Adoption}  
    \vspace{-2pt}
    \label{figure8_gender_difference}
    \includegraphics[width=1.\textwidth, height=0.6\textheight]{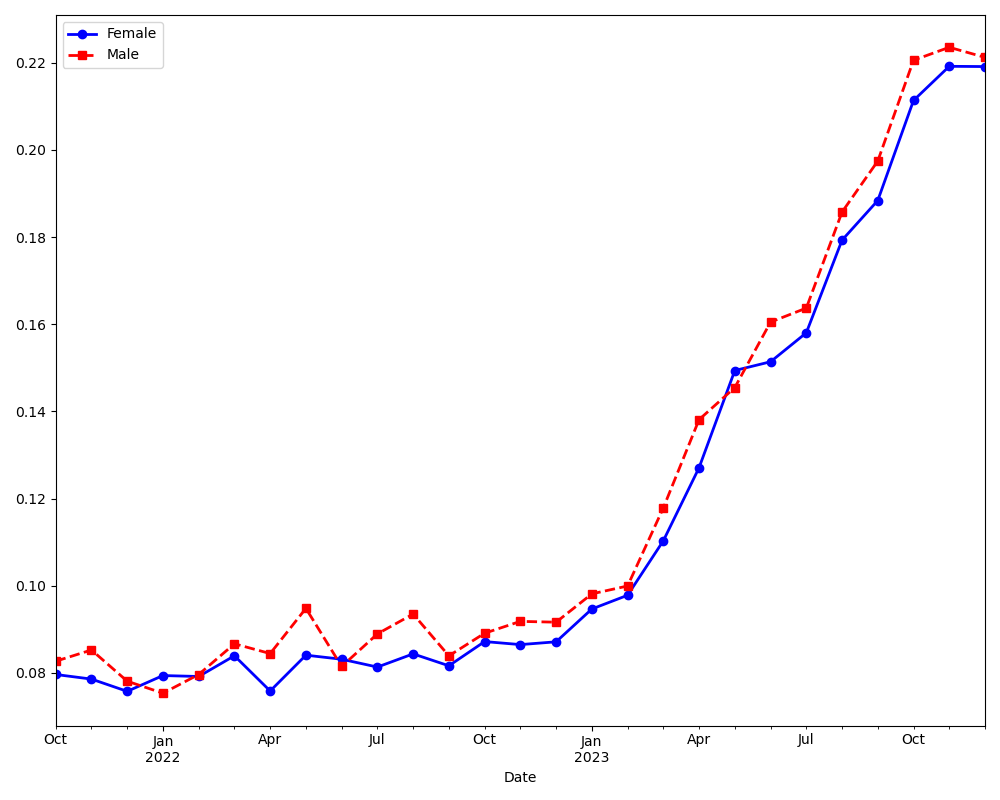}
    \vspace{-2pt}
\end{center}   
{\footnotesize 
Note: This figure examines gender differences in adopting GPT for revising articles. We first use a machine learning algorithm to estimate the gender of each author based on their name \footnote{Specifically, we use the “gender\_guesser” package in Python to classify authors as male, female, mostly male, mostly female, andy, or unknown. An author is classified as male if identified as male or mostly male, and female if identified as female or mostly female.}. Each month, we divide the sample into subgroups based on gender and calculate the percentage of abstracts revised by GPT. The figure shows no significant gender difference in GPT usage, as both male and female authors follow nearly the same trend in adopting GPT for revisions.
}   
\end{figure}

Figure \ref{figure8_gender_difference} illustrates gender differences in the adoption of GPT for abstract revisions. Interestingly, no significant differences are observed between male and female researchers regarding GPT adoption. Both genders exhibit similar trajectories in their adoption patterns over time, with an increasing reliance on GPT for revisions toward the end of 2023. 

\begin{figure}[H]
\begin{center}
   \caption{Seniorioty Difference in GPT Adoption}  
   \vspace{-2pt}
   \label{figure9_seniority_difference_in_adoption}
   
   \begin{subfigure}[t]{0.5\textwidth}
       \centering
       \includegraphics[width=\textwidth, height=0.5\textheight]{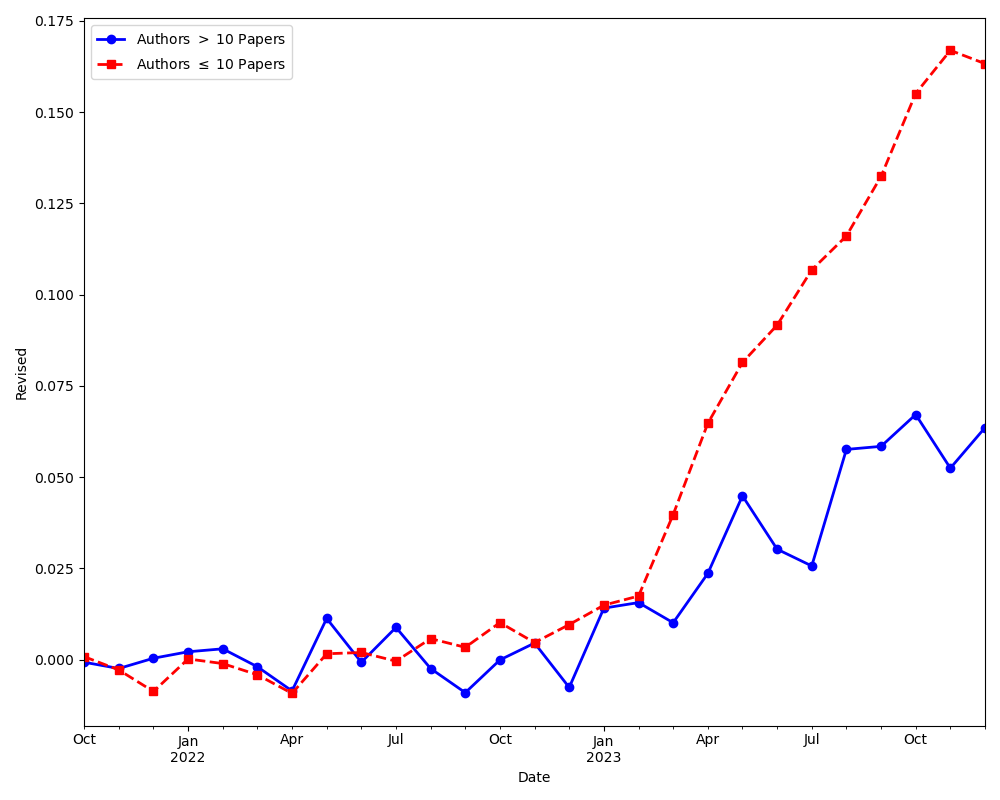}
       \caption{Panel A}
   \end{subfigure}
   \hfill
   \begin{subfigure}[t]{0.49\textwidth}
       \centering
       \includegraphics[width=\textwidth, height=0.5\textheight]{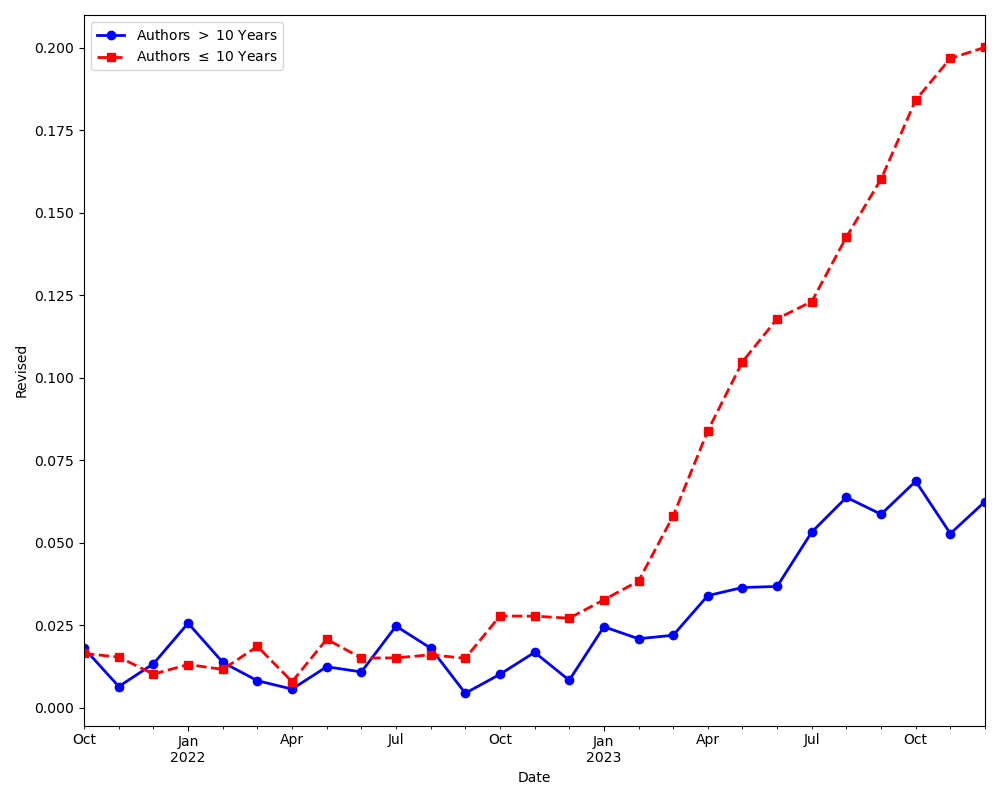}
       \caption{Panel B}
   \end{subfigure}
   
   \vspace{-2pt}
\end{center}   

{\footnotesize 
Note: This figure examines the differences in seniority in adopting GPT for revising articles. We use two measures as proxies for academic research experience: the number of academic papers written and years in academia. An author is classified as senior if they have written at least ten articles in our sample before 2021 or have been active in academia for at least ten years since their first paper was posted on arXiv. Each month, we divide the sample into subgroups based on author seniority and calculate the percentage of abstracts revised by GPT. The solid blue line represents the percentage of abstracts revised by GPT for the group where all authors have written at least ten papers. In comparison, the red dashed line represents the percentage for the group where all authors have written fewer than ten papers. The figure clearly shows that junior and senior authors followed a similar, non-trending pattern before introducing ChatGPT. However, junior authors became more active in using GPT to revise their articles after its introduction. Panel A measures seniority in terms of the number of papers written, while Panel B measures seniority based on years in academia.
}   
\end{figure}

Figure \ref{figure9_seniority_difference_in_adoption} examines GPT adoption for abstract revisions based on researchers' seniority. Two proxies for seniority are employed: the number of papers written and the duration of an academic career. Researchers are classified as senior if they have authored at least ten articles posted on arXiv before 2021 or have accumulated at least ten years of academic research experience before 2021. Researchers who do not meet these criteria are classified as junior. The results reveal notable differences in adoption patterns, with junior researchers demonstrating a higher propensity for using GPT.

The heterogeneous adoption of GPT across dimensions such as nativeness, ethnicity, gender, and academic experience underscores a significant selection effect in its use for academic revisions. Researchers with less experience or those less proficient in academic writing are more likely to adopt GPT, suggesting that it serves as an equalizing tool to bridge gaps in writing proficiency and academic experience.

Panels A and B of Figure \ref{figure9_seniority_difference_in_adoption} illustrate a notable trend in GPT adoption based on seniority, where seniority is defined by the number of academic papers authored or years of academic experience, respectively. The data reveal that junior researchers have significantly increased their use of GPT for abstract revisions, while senior researchers exhibit a relatively stable but lower adoption rate. By the end of 2023, the percentage of GPT adoption among junior researchers was nearly three times higher than that of senior researchers, highlighting the critical role GPT plays in supporting early-career academics in improving their writing.

Thus far, our exploration has not accounted for the potential correlations among variables. For example, native speakers may be highly correlated with certain ethnicities. To further investigate differences in GPT usage across various dimensions while controlling for other characteristics, we employ a multi-label logistic regression model:

\begin{equation}
\label{mul_logistic_reg}
P\left(y_{kt} = i \mid x_{kt} \right) = \frac{\exp \left(\beta_{i0} + x_{kt}^{\prime} \beta_i\right)}{1 + \sum_{j=1}^6 \exp \left(\beta_{j0} + x_{kt}^{\prime} \beta_j\right)}, \quad i = 1, 2, \ldots, 6,
\end{equation}
where \(P\left(y_{kt} = i \mid x_{kt} \right)\) represents the probability that abstract \(k\) at time \(t\) is revised following the \(i\)-th revision style. The unrevised version is treated as the baseline, with all coefficients set to 0. The vector \(x_{kt}\) includes observable characteristics of article \(k\), such as its academic field, the percentage of male and female authors, percentage of non-native writers, seniority, ethnicity, and the article's written date.

Table \ref{tab5_heterogeneity in the gpt adoption} presents the main results of the multivariable regression analysis. Due to missing information on the institutions for some articles, our sample size is reduced to 490,286 articles. First, in terms of disciplines, compared to the baseline (Mathematics), other fields are more likely to use GPT for comprehensive revisions rather than merely for grammar or syntax checks. All other disciplines consistently exhibit a higher likelihood than Mathematics to adopt comprehensive revisions, with the effect being particularly pronounced in Biology, Computer Science, and Electrical Engineering and System Science (EE\&SS). Except for Computer Science, all other disciplines are less inclined to use GPT for grammar checks or corrections (revision 5) or to improve writing objectivity than Mathematics.

In terms of ethnicity, researchers of East Asian and Muslim backgrounds are more likely to leverage GPT across all dimensions of revision, with the most significant effect seen in comprehensive revisions. In contrast, British researchers are more likely to use GPT for targeted revisions rather than for comprehensive changes. Regarding seniority, senior researchers rely less on GPT for revising their articles, particularly in terms of comprehensive revisions. Non-native speakers, such as researchers from Africa, East Asia, India, and Muslim countries, are more inclined to use GPT for comprehensive revisions relative to the baseline (Western European researchers).

\section{{Impact on} Writing Style}

The rapid adoption of large language models (LLMs) in scientific writing raises a critical question: what is the impact of GPT-assisted writing on writing style and quality? A substantial body of literature \cite{chiswick2004writing,weinberger2015ten,frassl2018ten,dejesus2019generic,ryba2019can,forero2020brief,ryba2021better,kousha2024factors} has explored various dimensions of scientific writing and its complex relationship with knowledge dissemination, publication, and citation. For instance, a recent study \cite{weinberger2015ten} conducted a systematic review and analysis of advances in scientific writing, presenting ten fundamental rules for crafting abstracts and keywords that significantly enhance the likelihood of publication and citation. The study focused on articles in Biology and highlighted key principles for improving textual quality.

Building on this work, we estimate the impact of GPT on writing by comparing the textual features of human-written and GPT-revised abstracts across ten dimensions. We exclude dimensions reliant on keyword information, which is unavailable in our dataset \footnote{Due to the absence of keyword information in our metadata, we limit our analysis to rules that do not require keyword-based evaluation.}.

Specifically, these rules include: 1) "Keep It Short," which we measure using the number of words (Rule 1a) and the number of sentences (Rule 1b); 2) "Keep It Compact," which encourages breaking compound sentences into simple ones and removing redundant words, measured by the percentage of sentences containing fewer than 20 words; 3) "Keep It Simple," which is typically measured by the percentage of words appearing in the Keywords (not measured due to missing Keywords); 4) "Use the Present Tense," measured as the ratio of present tense to the total of present and past tense verbs; 5) "Avoid Adjectives and Adverbs," discouraging the use of unnecessary modifiers, measured by the proportion of adjectives and adverbs to total words; 6) "Focus," which advises sticking to a single point, usually measured by the proportion of words appearing in the Keywords (also not measured due to missing Keywords); 7) "Signal Importance and Novelty," measured by the presence of words signaling novelty (Rule 7a) and importance (Rule 7b); 8) "Be Bold," encouraging strong assertions, measured as the ratio of superlatives to the total of superlatives and comparatives; 9) "Show Confidence," discouraging overuse of hedge words (e.g., “somewhat,” “speculative,” “appear,” “almost,” “largely”), measured by the count of hedge words; and 10) "Avoid Evocative Words," which discourages excessive use of highly evocative terms, measured by the counts of pleasant and unpleasant words (Rules 10a and 10b).

{\subsection{The potential impact of GPT on writing style}}

Our analysis consists of two essential components. First, we quantify the potential impact of LLMs on writing through a thought experiment that directly compares an original arXiv abstract with its GPT-revised counterpart based on a specific prompt. This comparison enables us to measure the differences between each abstract and its GPT-revised version while controlling for article-level features such as research field, topic, and content. One notable advantage of this approach is its ability to mitigate the challenges of selection bias, as the decision to use GPT for revisions reflects an optimal choice by the researcher. Consequently, GPT-revised abstracts may systematically differ from unrevised abstracts, as discussed in Section \ref{sec:gpt_adoption} on AI adoption.

To examine the differences between the original and revised abstracts for each article, we employ a linear regression model that accounts for paper-level characteristics and includes article fixed effects:

\begin{equation}
\label{heterogeneity_in_rule_bt_original_revise}
y_{it} = \beta_0 + \sum_{k=1}^6 \beta_k x_{ikt} + \alpha_i  + \epsilon_{it}
\end{equation}

Here, \(y_{it}\) represents the measured value for each rule of interest (e.g., number of words, number of sentences, etc.), and \(x_{ikt}\) is a dummy variable for article \(i\) at its latest update in month \(t\), equal to 1 if the abstract was revised following the \(k\)-th prompt, and 0 otherwise. The term \(\alpha_i\) captures article-specific fixed effects, enabling comparisons between an article's original and revised versions. \(\beta_0\) represents the baseline for the abstract. Since each article is associated with a unique update date, controlling for article fixed effects also controls for time-fixed effects. We test whether the coefficients \(\beta_k\) (\(k = 1, 2, \ldots, 6\)) significantly differ from zero and whether the revised versions adhere to the basic rules that promote publication and citations. 

Table \ref{tab5_panelA_Measure_the_Difference_of_original_and_GPT_revised_Text} presents results with paper fixed effects, while Table \ref{tab5_Measure the Difference of original and GPT-revised Texts} controls for paper-level characteristics. Our regression analysis reveals three key insights:
\begin{itemize}
    \item \textbf{Significant Differences}: Across all versions and rules, there are substantial differences between the original abstracts and their GPT-rewritten versions, as reflected in the standard deviations and means (see Table \ref{tab_app_summary_statistics}).
    
    \item \textbf{Conciseness and Clarity}: GPT-revised abstracts tend to be shorter, with fewer words and sentences. This indicates GPT's tendency to prioritize brevity and clarity. For example, prompts 1--5 lead to a reduction of over 25\% in word count and shorten more than one sentence on average, while the comprehensive revision reduces the abstract by more than 10\%.

    \item \textbf{Shift in Writing Style}: GPT revisions are less likely to emphasize novelty or use superlatives but more likely to highlight importance. Interestingly, the first five versions, each focusing on a different writing aspect, produce consistent changes. However, the final version, which integrates all aspects, results in a distinct style with more and longer sentences. This may be due to the challenge of balancing multiple writing considerations, which can inadvertently lengthen the text.

    \item \textbf{Tense and Word Usage}: All revisions are more likely to use the present tense rather than the past tense. The comprehensive revision is less likely to use adjectives and adverbs compared to the revisions following specific guidelines. Additionally, the revised versions consistently use fewer hedge words, with usage reduced by roughly one-third across all revisions. GPT-revised versions also avoid using pleasant or unpleasant words, signaling that GPT tends to adopt a more neutral tone.
\end{itemize}

To better understand the heterogeneity in writing rules across different dimensions, we replaced the article fixed effects with observable characteristics and date fixed effects. The results are summarized in Table \ref{tab5_Measure the Difference of original and GPT-revised Texts}. We included variables such as the ethnicity of the authors (measured as the percentage of Africans, British, East Asian, Western European authors, etc.), the discipline of the article, the gender distribution (male, female, and unknown), the average number of papers published by the authors, and the average number of academic years.

A notable finding is that GPT-revised abstracts—particularly those revised comprehensively under prompt 6—closely align with the writing style typically exhibited by senior authors, as measured by academic years. In 8 out of 11 writing rules, GPT mirrors the preferences of senior authors. For example, GPT tends to generate abstracts that feature fewer words, fewer sentences, fewer short sentences, a higher ratio of present tense to past tense, more frequent use of adjectives and adverbs, and less frequent use of novelty and emotionally charged words (both unpleasant and pleasant).

Overall, these findings suggest that GPT has a significant influence on scientific writing, though its effects depend heavily on the specific prompts and instructions provided. This highlights the potential of GPT to standardize academic writing styles while also emphasizing the importance of prompt design in shaping its outputs.

{\subsection{The potential impact of GPT on writing style}}

The differences documented in Tables \ref{tab5_panelA_Measure_the_Difference_of_original_and_GPT_revised_Text} and \ref{tab5_Measure the Difference of original and GPT-revised Texts} suggest that the use of LLMs may significantly enhance writing quality. This raises a critical question: is there any detectable real-world difference between manuscripts revised by LLMs and those that are not? To address this, our second analysis focuses on original abstracts posted on arXiv after November 2022, examining the writing features of those identified as LLM-revised. The following specification is used:

\begin{equation}
\label{heterogeneity_of_rule_bt_unrevised_revised}
y_{it} = \beta_0 + \sum_{k=1}^6 \beta_k x_{ikt} + z_{it}^{\prime} \gamma + \gamma_t + \epsilon_{it},
\end{equation}

where \(y_{it}\) represents the measurements of writing rules (e.g., word count, sentence length). The variable \(x_{ikt}\) equals 1 if the abstract has been revised according to version \(k\), and 0 otherwise. The term \(z_{it}\) includes control variables at the article level, and \(\gamma_t\) represents time-specific fixed effects.

Table \ref{tab7_Difference in the Writing Rules between Human and GPT-written} compares unrevised and revised abstracts identified by our fine-tuned models. A key observation is the significant selection bias associated with the use of specific prompts for revision, as documented in Table \ref{tab5_heterogeneity_in_gpt_adoption} on GPT adoption. This selection bias complicates the interpretation of results in Table \ref{tab7_Difference in the Writing Rules between Human and GPT-written}. For instance, researchers with less experience, non-native speakers, or those in fields like Computer Science and Biology are more likely to use GPT for comprehensive revisions. These groups also tend to write longer abstracts, which introduces endogeneity concerns into the regression analysis. Simple regression methods may therefore fail to fully account for the confounding effects stemming from this selection bias.

\subsection{Heterogeneous Convergence in Writing Style: Thought Experiment}
With the rapid adoption of AI in writing and measurable improvements in writing quality based on simple writing rules, a critical question emerges: Does following GPT's guidance lead to convergence in writing styles among researchers? In this section, we explore how adopting GPT influences writing style convergence across different researcher groups.

Our analysis proceeds from two perspectives. First, by examining the textual similarity between the original abstracts and their GPT-revised counterparts, we demonstrate significant heterogeneity in convergence across different groups of researchers. Second, we assess whether original abstracts written by different researcher groups exhibit convergence. Specifically, we denote the collection of abstracts as \(C_i\), where \(i = 0, 1, 2, \ldots, 6\). Here, \(C_0\) represents original abstracts from arXiv, while \(C_i\) corresponds to abstracts revised according to a specific GPT prompt \(i\).

To quantify similarity, we define a term distribution over the vocabulary for each abstract, represented as \(\mathbf{v}_d\), which is \(L_2\)-normalized. We then calculate the cosine similarity between two abstracts, \(d_1\) and \(d_2\), as follows:
\[
\text{Cosine Similarity}(d_1, d_2) = \frac{\mathbf{v}_{d_1} \cdot \mathbf{v}_{d_2}}{\|\mathbf{v}_{d_1}\| \|\mathbf{v}_{d_2}\|}
\]

This metric allows us to quantify the similarity between original and GPT-revised abstracts, as well as across researcher groups. By analyzing these patterns, we gain insights into how GPT adoption influences convergence in writing styles, revealing nuanced effects across different researcher demographics.

Our main findings are twofold. First, we calculate the similarity between the original abstracts on arXiv and their GPT-revised counterparts. For each month, we average the similarity across all articles, focusing on the evolution of the similarity between the original abstracts and those revised using the comprehensive prompt.

\begin{figure}[H]
\begin{center}
   \caption{Heterogeneous Impact of GPT Adoption (Seniority)}  
   \vspace{-2pt}
   \label{figure10_similarity_by_seniority}
   
   \centering
   \includegraphics[width=0.7\textwidth, height=0.5\textheight]{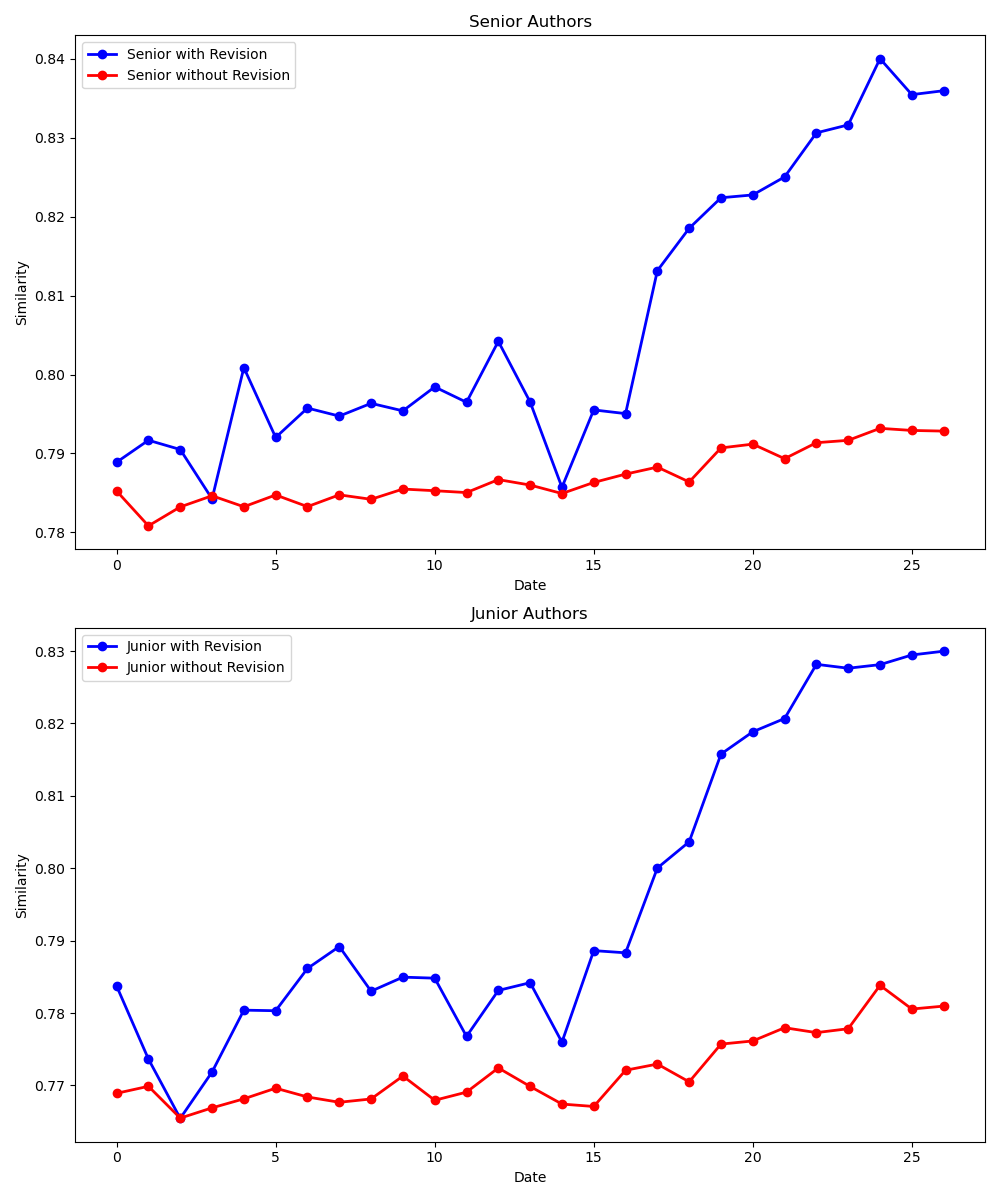}
   \caption*{Panel A: All Authors are Seniors}
   
   \vspace{-2pt}
\end{center}   
{\footnotesize 
Note: This figure shows the similarity in writing between the abstract and its GPT-revised version for junior and senior researchers. The top subplot shows the textual similarity for the senior authors while the bottom shows that for the junior counterparts. The red line shows the similarity between the abstract and its GPT-revised version for seniors without using GPT, and the blue line shows the similarity for seniors adopting GPT. The bottom depicts display similar analysis.  For each article, we compare the abstract with its GPT-revised version (version 6) by constructing a bag-of-words representation and calculating the cosine similarity between the two versions. We compute the average cosine similarity for articles authored by senior and junior researchers separately each month. Authors are considered senior they have at least ten years of research experience; otherwise, they are classified as junior. In Panel A, a paper is considered written by seniors if at least one author is senior (Similar results hold if we define seniority based on the number of papers written).
}   
\end{figure}

Figure \ref{figure10_similarity_by_seniority} illustrates the heterogeneous convergence in writing styles for articles authored by junior and senior researchers. We define junior authors as those with less than ten years of academic research experience. Panel A examines the convergence for senior authors, while Panel B focuses on junior authors.

Several key observations emerge. First, for both junior and senior authors, the textual similarity between the original abstract and its GPT-revised counterpart remains stable prior to the public release of ChatGPT. Second, following the release of ChatGPT, the writing style increasingly converges to that of the GPT-revised abstracts, irrespective of author seniority. However, this convergence is limited to junior and senior authors who actively use GPT for article revisions. For authors who do not adopt GPT, the textual similarity remains stable over time. 

\begin{figure}[H]
\begin{center}
   \caption{Heterogeneous Impact of GPT-Adoption (Gender/Native)}  
   \vspace{-2pt}
   \label{figure12_similarity_bt_native_non_native}
   \begin{subfigure}[t]{0.49\textwidth}
       \centering
       \includegraphics[width=\textwidth, height=0.5\textheight]{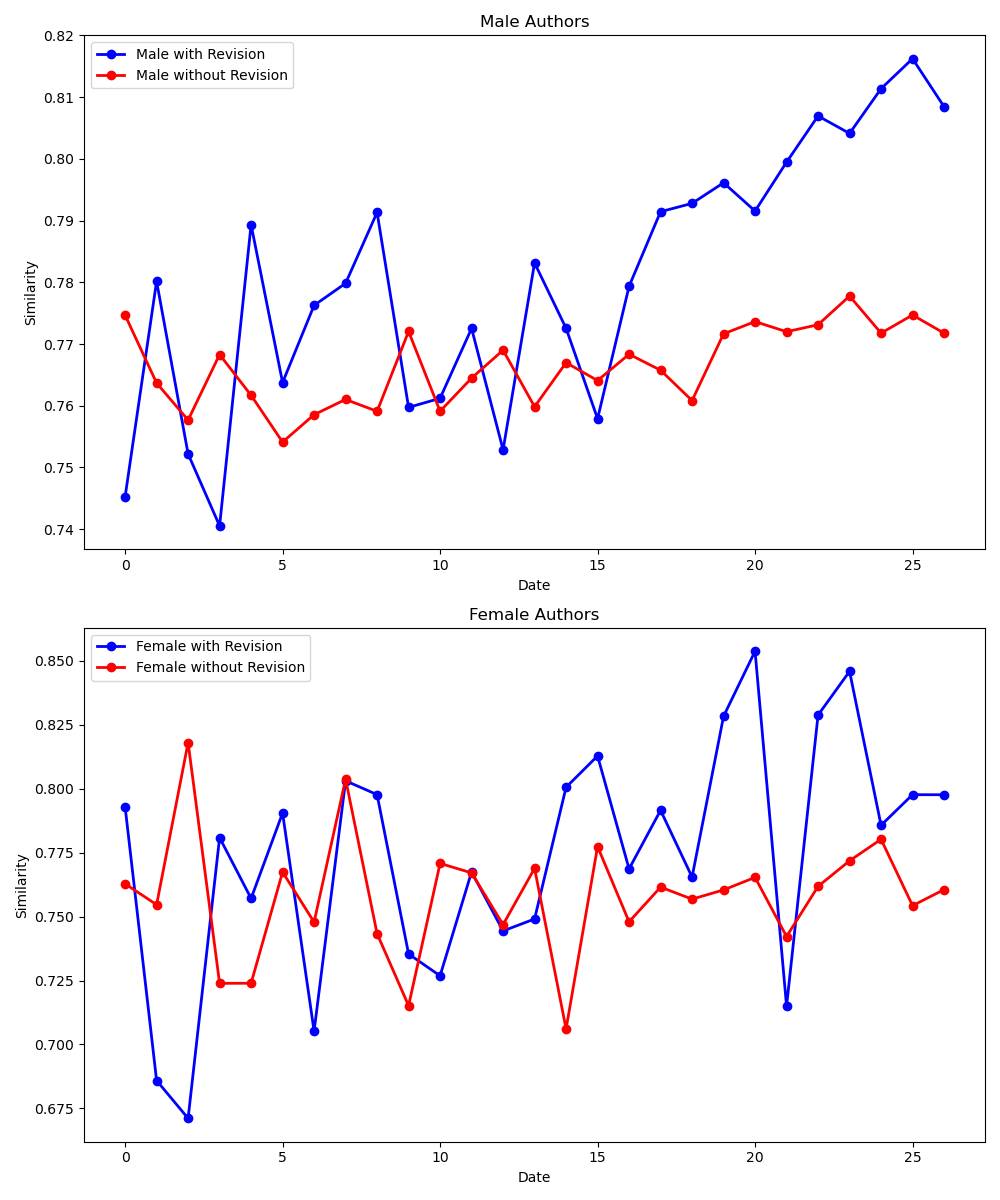}
       \caption*{Panel A: Female and Male Authors}
   \end{subfigure}
   \hfill
   \begin{subfigure}[t]{0.49\textwidth}
       \centering
       \includegraphics[width=\textwidth, height=0.5\textheight]{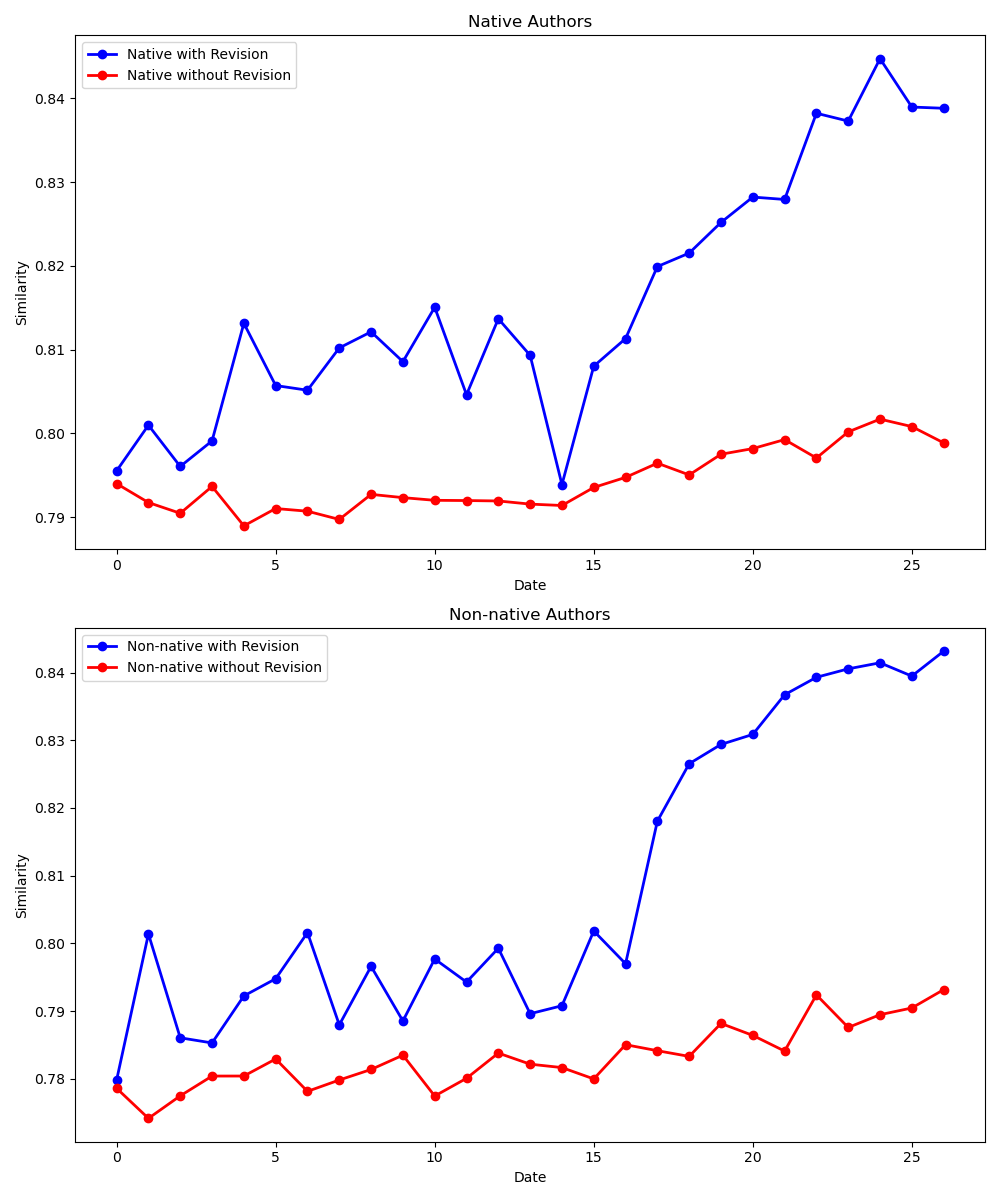}
       \caption*{Panel B: Native and Non-native Authors}
   \end{subfigure}
   \vspace{-2pt}
\end{center}   

{\footnotesize 
Note: This figure illustrates the textual similarity between abstracts and their GPT-revised versions across different dimensions. Panel A highlights the gender differences in GPT adoption. In the top subplot, the blue line represents the textual similarity for male authors adopting GPT, while the red line shows the similarity for males without GPT adoption. The bottom subplot presents a similar analysis for female authors. Panel B focuses on the differences in GPT adoption between native and non-native authors. The top subplot shows the results for native authors, with the blue line representing those adopting GPT and the red line depicting those not using GPT. The bottom subplot mirrors this analysis for non-native authors. For each article, we construct a bag-of-words representation and calculate the cosine similarity between the original and revised versions, averaging the results monthly.
}   
\end{figure}

Similarly, Figure \ref{figure12_similarity_bt_native_non_native} illustrates the heterogeneous convergence between gender and nativeness. Panel A displays the convergence patterns for male and female authors, while panel B shows the convergence for native and nonnative authors. 

For gender, an interesting pattern emerges: only male researchers who adopt GPT exhibit convergence to GPT-style writing, while female authors do not show a significant convergence trend. Regarding nativeness, both native and nonnative authors who use GPT display a clear convergence to GPT-style writing after the public release of ChatGPT. In contrast, authors who do not adopt GPT maintain a stable level of similarity with GPT revisions. 

\subsection{Heterogeneous Convergence in Writing Style: Real World}
The previous subsection analyzed heterogeneous convergence to GPT's writing style by comparing original abstracts to their GPT-revised counterparts across different researcher groups. This section shifts the focus to whether similar convergence patterns can be observed in actual written texts, examining the heterogeneity by seniority, discipline, gender, and nativeness.

Figure \ref{figure11_similarity_bt_junior_senior} explores convergence patterns based on seniority and discipline. The Left Panel assesses whether juniors' writing styles converge with those of senior researchers. Two key findings emerge. First, regardless of whether seniority is measured by academic years or the number of papers published, juniors who use GPT for revision exhibit a significant convergence toward senior writing styles after the public release of ChatGPT. However, this convergence does not occur for juniors who do not adopt GPT. A plausible explanation is that GPT-revised articles resemble those written by more experienced authors, as shown in Table \ref{tab5_panelA_Measure_the_Difference_of_original_and_GPT_revised_Text}. Thus, leveraging GPT for revisions may push juniors' writing styles closer to those of senior authors. Second, while juniors using GPT become more similar to seniors, their writing remains less similar to that of seniors compared to juniors who do not use GPT. One possible reason is selection bias: juniors with stronger writing skills may opt not to use GPT, while those with weaker skills may rely on it for revisions.

An additional point of interest is the heterogeneity in convergence across different academic disciplines. For instance, intuitively, fields like mathematics may rely less on writing style and more on results. The Right Panel of Figure \ref{figure11_similarity_bt_junior_senior} shows the similarity between juniors and seniors who adopt GPT, broken down by discipline. There is considerable variation across disciplines. After the release of ChatGPT, juniors from Electrical Engineering \& Computer Science (EE\&CS), Biology, Economics \& Finance, and Statistics show a marked convergence toward senior writing styles. In contrast, juniors in Mathematics exhibit little to no increase in similarity to their senior counterparts' writing styles.

Figure \ref{figure12_similarity_bt_native_non_native} illustrates the heterogeneous convergence along gender and nativeness. The left panel presents results for gender, while the right panel focuses on nativeness. For the gender analysis, we provide two sets of results. The top subplot shows the similarity between male and female researchers, conditional on whether GPT is used or not. The bottom subplot highlights the similarity between researchers with and without GPT adoption, conditional on gender. Several key points emerge from the analysis.

First, conditional on whether researchers use GPT, we observe that the textual similarity between male and female researchers remains stable for those not adopting GPT. However, for those using GPT for revisions, there is a significant increase in similarity after the public release of ChatGPT. This suggests that the convergence in writing styles is indeed driven by GPT adoption. 

Second, conditional on gender, we find that only among male researchers does the textual similarity between those who use GPT and those who do not converge. Interestingly, this convergence is not observed for female researchers.

The right panel presents the writing style convergence between native and non-native researchers. Here, we observe an interesting but distinct pattern. First, conditional on GPT usage, writing style convergence between native and non-native researchers occurs only for those who used GPT for academic revisions. Among those not using GPT, there is no convergence between native and non-native researchers. 

Second, focusing on native and non-native researchers, we find that the writing styles of those who adopt GPT and those who do not tend to converge. A possible explanation lies in a selection effect: researchers with less experience in writing are more likely to use GPT for polishing their work. Since GPT revisions are often more similar to those of experienced writers, this leads to a convergence between researchers who use GPT and those who do not.

\section{Conclusion}\label{conc}
This study develops a systematic framework and provides a comprehensive analysis of GPT usage in academic writing, examining its impact on writing quality, style, and convergence across different researcher groups. Our findings reveal that GPT significantly enhances writing quality by improving clarity, conciseness, and professionalism. However, these enhancements are accompanied by a notable shift toward a more formal and neutral tone, aligning GPT-revised texts with the structured writing styles of senior researchers. This shift suggests that while GPT improves textual quality, it may also contribute to the standardization of writing practices, potentially reducing the diversity of academic expression.

Our analysis also highlights substantial convergence in writing styles across researcher groups following GPT adoption. Junior researchers, in particular, demonstrate a pronounced shift toward writing styles that mirror those of more experienced scholars. Additionally, male and non-native authors exhibit higher levels of convergence in writing styles after adopting GPT, whereas female authors show less significant changes. Importantly, these patterns are primarily driven by researchers actively using GPT for writing revisions. These findings suggest that GPT has the potential to democratize academic writing by elevating the quality of less experienced researchers’ work, yet it also raises concerns about the homogenization of writing styles among those who heavily rely on the technology.

This study opens several avenues for future research. First, a deeper investigation into the long-term effects of GPT on academic writing diversity is needed, particularly as LLMs continue to evolve and integrate more deeply into the academic workflow. Second, future studies could examine how GPT impacts broader aspects of scientific communication, such as the development of ideas, argumentation, or originality, beyond technical improvements in writing style. Finally, exploring the ethical implications of GPT usage in academia—particularly regarding issues of authorship, originality, and accountability—will be essential as AI becomes increasingly entrenched in scientific research.

\setlength{\bibsep}{4.0pt}
\bibliographystyle{plainnat}

\clearpage
\bibliography{reference.bib}

\clearpage
\section{Figures and Tables}

\begin{table}[H]
\centering
\caption{The Number of Articles by Field and Year in arVix}
\label{tab1_number_of_articles}
\parbox[t]{5.5in}{\footnotesize{  This table shows the number of articles by field and year. We abbreviate Mathematics as “Maths”, Physics as “Phys”, Computer Science as “CS”, Electrical Engineering and System Science as “EE\&SS”, Statistics as “Stats”, Biology as “Bio”, Economics as “Econ”, and Finance as “Fin”.
    \\}
}
\begin{center}
\begin{tabular}{lrrrrrrrrr}
\hline
\hline
Year   & Maths  & Phys   &   CS   & EE\&SS & Stats  & Bio   & Econ  & Fin   & Total   \\
\hline
2021   & 36280  & 69824  & 60467  & 8185   & 5103   & 2187  & 1251  & 931   & 184228  \\
2022   & 38578  & 69928  & 65631  & 8970   & 5233   & 1979  & 1331  & 874   & 192524  \\
2023   & 49210  & 84966  & 92830  & 10731  & 7028   & 2655  & 2054  & 1158  & 250632  \\
\hline
Total  & 124068 & 224718 & 218928 & 27886  & 17364  & 6821  & 4636  & 2963  & 627384  \\
\hline
\hline
\end{tabular}
\end{center}
\end{table}
\clearpage

\begin{table}[H]
\centering
\caption{Model Performance: Training and Test Samples}
\label{tab2_out of sample performance evaluation}
\parbox[t]{5.5in}{\footnotesize{This table reports the model performance on both the training and test samples. The training sample includes all articles with their latest update between January 1, 2021, and September 31, 2021, while the test sample consists of articles updated between October 1, 2021, and November 31, 2021. The first column corresponds to a deep-learning classification model used to differentiate the original and revised abstracts in terms of 'Clarity and Conciseness.' The second through sixth columns represent classification models aimed at distinguishing the original abstracts from the revised versions for 'Formality' through 'Comprehensive Revision.' We report the predicted precision, recall, accuracy, and F1 score for both the training sample (Panel A) and the test sample (Panel B). The definitions of the metrics are as follows: Precision is the percentage of paragraphs identified as revised that were actually revised by AI. Recall refers to the percentage of documents that were actually revised and correctly identified as such by the model. Accuracy measures the overall percentage of paragraphs that the model classified correctly. The F1 score is the harmonic mean of precision and recall.
    \\}
}
\begin{subtable}{\textwidth}
\centering
\caption*{Panel A: The training sample}
\begin{tabular}{lcccccc}
\hline
\hline
 & Revision 1 & Revision 2 & Revision 3 & Revision 4 & Revision 5 & Revision 6 \\
\hline
No. of doc   & 273,490 & 273,490 & 273,490 & 273,490 & 273,490 & 273,490 \\
Precision    & 0.981   & 0.994   & 0.988   & 0.992   & 0.965   & 0.992   \\
Recall       & 0.987   & 0.998   & 0.996   & 0.997   & 0.973   & 0.995   \\
Accuracy     & 0.984   & 0.996   & 0.992   & 0.994   & 0.969   & 0.993   \\
F1\_score    & 0.984   & 0.996   & 0.992   & 0.994   & 0.969   & 0.993   \\
\hline
\hline
\end{tabular}
\end{subtable}

\vspace{12pt} 

\begin{subtable}{\textwidth}
\centering
\caption*{Panel B: The test sample}
\begin{tabular}{lcccccc}
\hline
\hline
 & Revision 1 & Revision 2 & Revision 3 & Revision 4 & Revision 5 & Revision 6 \\
\hline
No. of doc   & 63,904 & 63,904 & 63,904 & 63,904 & 63,904 & 63,904 \\
Precision    & 0.960  & 0.979  & 0.979  & 0.975  & 0.942  & 0.983  \\
Recall       & 0.975  & 0.991  & 0.991  & 0.989  & 0.962  & 0.989  \\
Accuracy     & 0.967  & 0.985  & 0.985  & 0.982  & 0.951  & 0.986  \\
F1\_score    & 0.967  & 0.985  & 0.985  & 0.982  & 0.952  & 0.986  \\
\hline
\hline
\end{tabular}
\end{subtable}
\end{table}
\clearpage

\begin{tiny}
\begin{longtable}{|>{\RaggedRight}p{15cm}|}
\caption{An Example of GPT Revised Texts}
\label{tab3:differnce-bt-ai-human} \\
\hline
\textbf{Original Abstract} \\
\hline
"Adaptive Moment Estimation (Adam), which combines Adaptive Learning Rate and Momentum, would be the most popular stochastic optimizer for accelerating the training of deep neural networks. However, it is empirically known that Adam often generalizes worse than Stochastic Gradient Descent (SGD). The purpose of this paper is to unveil the mystery of this behavior in the diffusion theoretical framework. Specifically, we disentangle the effects of Adaptive Learning Rate and Momentum of the Adam dynamics on saddle-point escaping and flat minima selection. We prove that Adaptive Learning Rate can escape saddle points efficiently, but cannot select flat minima as SGD does. In contrast, Momentum provides a drift effect to help the training process pass through saddle points, and almost does not affect flat minima selection. This partly explains why SGD (with Momentum) generalizes better, while Adam generalizes worse but converges faster. Furthermore, motivated by the analysis, we design a novel adaptive optimization framework named Adaptive Inertia, which uses parameter-wise adaptive inertia to accelerate the training and provably favors flat minima as well as SGD. Our extensive experiments demonstrate that the proposed adaptive inertia method can generalize significantly better than SGD and conventional adaptive gradient methods." \\
\hline
\textbf{Clarity and Conciseness} \\
\hline
"Adaptive Moment Estimation (Adam), which merges Adaptive Learning Rate and Momentum, is a leading stochastic optimizer for speeding up deep neural network training. It is empirically observed that Adam generally underperforms Stochastic Gradient Descent (SGD) in terms of generalization. This paper investigates the reasons behind this behavior within a diffusion theoretical framework. We analyze how Adaptive Learning Rate and Momentum affect Adam's ability to escape saddle points and select flat minima. Our findings confirm that while Adaptive Learning Rate efficiently escapes saddle points, it does not favor flat minima selection as SGD does. Conversely, Momentum aids in navigating through saddle points but barely influences minima selection. This sheds light on why SGD, especially with Momentum, generally achieves better generalization than Adam. Additionally, inspired by these insights, we introduce a new adaptive optimization strategy called Adaptive Inertia. This method uses adaptive inertia for each parameter to enhance training speed and effectively select flat minima akin to SGD. Our comprehensive experiments show that Adaptive Inertia outperforms both SGD and traditional adaptive gradient methods in generalization." \\
\hline
\textbf{Formality and Professionalism} \\
\hline
"Adaptive Moment Estimation (Adam), integrating Adaptive Learning Rate and Momentum, stands as a premier stochastic optimizer for expediting deep neural network training. Notably, Adam is empirically understood to generalize less effectively than Stochastic Gradient Descent (SGD). This study elucidates the underlying mechanisms of this phenomenon within a diffusion theoretical framework. We meticulously examine the influence of Adaptive Learning Rate and Momentum on Adam's dynamics concerning saddle point evasion and flat minima selection. It is demonstrated that although Adaptive Learning Rate efficiently circumvents saddle points, it fails to prioritize flat minima as effectively as SGD. In contrast, Momentum facilitates overcoming saddle points while minimally impacting the selection of flat minima. These findings partially elucidate the superior generalization performance of SGD enhanced with Momentum over Adam. Moreover, prompted by these analyses, we devise an innovative adaptive optimization model termed Adaptive Inertia, employing parameter-specific adaptive inertia to expedite training and favor flat minima comparably to SGD. Extensive experimental results validate that Adaptive Inertia surpasses both SGD and conventional adaptive gradient methods in terms of generalization."\\
\hline
\textbf{Objectivity} \\
\hline
"Adaptive Moment Estimation (Adam), combining Adaptive Learning Rate and Momentum, is frequently employed as a stochastic optimizer for accelerating deep neural network training. Research indicates that Adam typically exhibits inferior generalization compared to Stochastic Gradient Descent (SGD). This paper aims to clarify this disparity through a diffusion theoretical approach, analyzing the roles of Adaptive Learning Rate and Momentum in Adam's dynamics, particularly in escaping saddle points and selecting flat minima. Evidence suggests that although Adaptive Learning Rate effectively escapes saddle points, it does not select flat minima with the efficacy of SGD. Conversely, Momentum assists in bypassing saddle points without significantly affecting minima selection. This contributes to understanding why SGD, particularly with Momentum, outperforms Adam in generalization. Building on this foundation, we propose an innovative optimization method called Adaptive Inertia, which leverages parameter-specific adaptive inertia to accelerate training and select flat minima as effectively as SGD. Empirical tests confirm that Adaptive Inertia provides superior generalization compared to SGD and traditional adaptive gradient approaches."
\\
\hline
\textbf{Readability and Understandability} \\
\hline
"Adaptive Moment Estimation (Adam), which blends Adaptive Learning Rate with Momentum, is widely used to speed up training deep neural networks. It's known that Adam doesn't generalize as well as Stochastic Gradient Descent (SGD). This paper explores why this is the case using a diffusion theory framework. We look at how Adaptive Learning Rate and Momentum affect Adam's ability to move past saddle points and select flat minima. We find that Adaptive Learning Rate gets past saddle points well but doesn't pick flat minima like SGD. Meanwhile, Momentum helps get past saddle points but doesn't really impact minima selection. This helps explain why SGD with Momentum tends to generalize better than Adam. Also, based on our findings, we've developed a new optimization method called Adaptive Inertia. This method uses adaptive inertia tailored to each parameter to speed up training and select flat minima as effectively as SGD. Tests show that Adaptive Inertia generalizes better than both SGD and other adaptive gradient methods."
 \\
\hline
\textbf{Grammar and Syntax} \\
\hline
"Adaptive Moment Estimation (Adam), which integrates Adaptive Learning Rate with Momentum, is a prominent stochastic optimizer used to accelerate the training of deep neural networks. It is empirically evident that Adam typically generalizes less effectively than Stochastic Gradient Descent (SGD). This paper investigates this phenomenon within a diffusion theoretical framework. We analyze how Adaptive Learning Rate and Momentum influence Adam's dynamics, especially in escaping saddle points and selecting flat minima. Our findings confirm that while Adaptive Learning Rate efficiently escapes saddle points, it is less effective at selecting flat minima compared to SGD. In contrast, Momentum aids in navigating through saddle points but does not significantly impact minima selection. This partly explains why SGD, particularly with Momentum, tends to generalize better than Adam. Additionally, inspired by our analysis, we have developed a new adaptive optimization approach called Adaptive Inertia, which employs parameter-specific adaptive inertia to accelerate training and favor flat minima similarly to SGD. Our extensive experiments demonstrate that Adaptive Inertia generalizes significantly better than both SGD and conventional adaptive gradient methods."
\\
\hline
\textbf{Comprehensive Revision} \\
\hline
"Adaptive Moment Estimation (Adam), combining Adaptive Learning Rate with Momentum, is a premier stochastic optimizer for accelerating deep neural network training. Despite its popularity, Adam often generalizes less effectively than Stochastic Gradient Descent (SGD). This study delves into this issue within a diffusion theoretical framework, scrutinizing the impact of Adaptive Learning Rate and Momentum on Adam's ability to escape saddle points and select flat minima. We find that while Adaptive Learning Rate efficiently escapes saddle points, it falls short in selecting flat minima compared to SGD. Conversely, Momentum aids in navigating saddle points but scarcely affects minima selection. These insights partially elucidate why SGD, especially with Momentum, generally achieves better generalization than Adam. Furthermore, inspired by these findings, we introduce a novel optimization strategy named Adaptive Inertia. This approach leverages parameter-specific adaptive inertia to both accelerate training and effectively select flat minima akin to SGD. Our comprehensive experimental validation shows that Adaptive Inertia substantially outperforms SGD and traditional adaptive gradient methods in generalization."
\\
\hline
\end{longtable}
\end{tiny}
\clearpage

\begin{table}[H]
\centering
\caption{Confusion Matrix of Multi-classes Classification}
    \parbox[t]{5.5in}{\footnotesize{ This table shows the confusion matrix for the multi-classes classification model for the test sample. Specifically, the $(i,j)$ element shows the probability of the revision $i$ being classified as revision $j$ where $0$ indicate the original unrevised version.
    \\}
}
\label{tab4_confusion_matrix_multiclasses}
\begin{tabular}{llllllll}
\hline
\hline
 & 0 & 1 & 2 & 3 & 4 & 5 & 6 \\
\midrule
\hline
0 & 92.25\% & 4.23\% & 0.40\% & 0.37\% & 0.15\% & 2.20\% & 0.40\% \\
1 & 1.23\% & 68.39\% & 7.53\% & 7.18\% & 7.44\% & 4.65\% & 3.57\% \\
2 & 0.20\% & 15.01\% & 54.16\% & 14.37\% & 7.28\% & 4.59\% & 4.39\% \\
3 & 0.14\% & 9.45\% & 16.21\% & 51.81\% & 10.20\% & 6.80\% & 5.39\% \\
4 & 0.23\% & 20.28\% & 13.78\% & 20.77\% & 23.76\% & 13.67\% & 7.52\% \\
5 & 2.64\% & 22.76\% & 11.47\% & 18.62\% & 18.46\% & 18.28\% & 7.77\% \\
6 & 0.25\% & 8.93\% & 7.17\% & 6.53\% & 7.09\% & 6.88\% & 63.14\% \\
\hline
\hline
\bottomrule
\end{tabular}
\end{table}

\begin{table}[htbp]
\centering
\caption{The Heterogeneity in the Adoption of GPT}
\label{tab5_heterogeneity_in_gpt_adoption}
\parbox[t]{\textwidth}{\footnotesize{
This table presents the heterogeneous adoption of GPT across various prompts, estimated using a multiclass logistic regression model as specified in Equation \ref{mul_logistic_reg}. All reported errors are robust to heteroskedasticity. Coefficients marked with *, **, and *** indicate significance at the 10\%, 5\%, and 1\% levels, respectively.
}}
\scalebox{0.8}{%
\begin{tabular}{lcccccc}
\hline
\hline
\textbf{Variables} & Version 1 & Version 2 & Version 3 & Version 4 & Version 5 & Version 6 \\
\hline
\textbf{Africans} & 0.193*** & 0.351*** & 0.460** & 0.189 & 0.117 & 0.361*** \\
& (0.061) & (0.133) & (0.180) & (0.231) & (0.088) & (0.116) \\
\textbf{British} & 0.473*** & 0.019 & 0.406*** & 0.332*** & 0.199*** & -0.157** \\
& (0.029) & (0.073) & (0.090) & (0.117) & (0.043) & (0.067) \\
\textbf{EastAsian} & 0.386*** & 0.730*** & 0.275*** & 0.740*** & 0.343*** & 0.837*** \\
& (0.026) & (0.059) & (0.081) & (0.101) & (0.039) & (0.053) \\
\textbf{EastEuropean} & 0.052 & -0.106 & 0.301** & -0.084 & -0.063 & -0.354*** \\
& (0.043) & (0.109) & (0.123) & (0.180) & (0.063) & (0.106) \\
\textbf{Indian} & 0.166*** & 0.169** & -0.003 & 0.222* & -0.004 & 0.222*** \\
& (0.034) & (0.076) & (0.108) & (0.129) & (0.050) & (0.067) \\
\textbf{Jewish} & 0.319*** & 0.119 & 0.026 & 0.117 & 0.118* & -0.063 \\
& (0.045) & (0.112) & (0.151) & (0.186) & (0.067) & (0.105) \\
\textbf{Muslim} & 0.270*** & 0.571*** & 0.421*** & 0.770*** & 0.159*** & 0.807*** \\
& (0.042) & (0.087) & (0.131) & (0.138) & (0.061) & (0.072) \\
\textbf{Biology} & 0.357*** & 0.923*** & -0.279 & -0.111 & -2.126*** & 3.344*** \\
& (0.060) & (0.115) & (0.213) & (0.285) & (0.260) & (0.075) \\
\textbf{Computer Science} & 0.423*** & 0.716*** & -0.604*** & 0.979*** & 0.559*** & 2.102*** \\
& (0.019) & (0.044) & (0.060) & (0.068) & (0.024) & (0.059) \\
\textbf{Economics} & 0.797*** & 0.691*** & 0.185 & -1.754** & -1.584*** & 1.087*** \\
& (0.057) & (0.141) & (0.195) & (0.710) & (0.225) & (0.166) \\
\textbf{Electrical Eng.} & 0.427*** & 1.463*** & -0.282*** & 0.667*** & -1.405*** & 1.933*** \\
& (0.031) & (0.054) & (0.108) & (0.108) & (0.088) & (0.070) \\
\textbf{Finance} & 1.261*** & 0.933*** & -1.170** & -1.812* & -1.324*** & 2.091*** \\
& (0.062) & (0.167) & (0.502) & (1.002) & (0.269) & (0.143) \\
\textbf{Physics} & 0.212*** & 0.263*** & 0.177*** & -0.184** & -0.329*** & 1.013*** \\
& (0.019) & (0.047) & (0.051) & (0.081) & (0.028) & (0.063) \\
\textbf{Statistics} & 0.235*** & 0.688*** & -0.647*** & -3.015*** & -0.769*** & 1.616*** \\
& (0.039) & (0.081) & (0.155) & (0.709) & (0.081) & (0.086) \\
\textbf{Female} & 0.113*** & -0.358*** & -0.462*** & 0.040 & 0.062 & -0.124* \\
& (0.035) & (0.082) & (0.116) & (0.131) & (0.052) & (0.069) \\
\textbf{Male} & -0.018 & -0.330*** & -0.263*** & -0.064 & 0.048 & -0.340*** \\
& (0.021) & (0.047) & (0.062) & (0.080) & (0.032) & (0.043) \\
\textbf{Paper Seniority} & -0.002 & -0.271*** & -0.385*** & -0.759*** & -0.144*** & -0.691*** \\
& (0.026) & (0.057) & (0.082) & (0.098) & (0.037) & (0.052) \\
\textbf{Year Seniority} & -0.234*** & -0.523*** & 0.016 & 0.028 & -0.297*** & -0.625*** \\
& (0.028) & (0.065) & (0.087) & (0.108) & (0.042) & (0.062) \\
\textbf{Obs.} & 490,286 & 490,286 & 490,286 & 490,286 & 490,286 & 490,286 \\
\hline
\hline
\end{tabular}%
}
\end{table}

\begin{table}[htbp]
\centering
\caption{Writing Rules: Within Paper Comparison}
\label{tab5_panelA_Measure_the_Difference_of_original_and_GPT_revised_Text}
\parbox[t]{\textwidth}{\footnotesize{
This table shows the differences in writing rules between the original abstracts on arXiv and their revised versions following specific prompts, estimated using regression \ref{heterogeneity_in_rule_bt_original_revise}. Article-level fixed effects are included to allow for a direct comparison within the same article. Year-month fixed effects account for time-varying changes in writing style. Rules 1a and 1b measure the number of words and sentences, rule 2 counts short sentences (fewer than 20 words), rule 4 captures the present-to-past tense ratio, rule 5 measures the proportion of adjectives and adverbs, rule 7a and 7b track novelty and importance keywords, rule 8 evaluates superlative-to-comparative ratios, rule 9 counts hedge words, and rules 10a and 10b measure the occurrence of pleasant and unpleasant words. Errors are robust. Coefficients marked *, **, and *** are significant at the 10\%, 5\%, and 1\% levels.
}}
\begin{adjustbox}{center, max width=0.95\textwidth}
\scalebox{0.75}{%
\begin{tabular}{lccccccccccc}
\hline
\hline
& \textbf{Rule1a} & \textbf{Rule1b} & \textbf{Rule2} & \textbf{Rule4} & \textbf{Rule5} & \textbf{Rule7a} & \textbf{Rule7b} & \textbf{Rule8} & \textbf{Rule9} & \textbf{Rule10a} & \textbf{Rule10b} \\
\hline
\textbf{Version1} & -46.44*** & -1.290*** & 4.036*** & 4.830*** & 1.620*** & -5.546*** & 8.502*** & -4.937*** & -2.608*** & -1.080*** & -0.918*** \\
& (0.0685) & (0.00266) & (0.0275) & (0.0160) & (0.00342) & (0.0335) & (0.0538) & (0.0310) & (0.0227) & (0.0159) & (0.0123) \\
\textbf{Version2} & -45.06*** & -1.346*** & 1.405*** & 3.849*** & 1.437*** & -5.313*** & 5.300*** & -5.473*** & -2.781*** & -1.197*** & -0.956*** \\
& (0.0695) & (0.00269) & (0.0273) & (0.0169) & (0.00345) & (0.0354) & (0.0540) & (0.0322) & (0.0232) & (0.0164) & (0.0125) \\
\textbf{Version3} & -46.95*** & -1.466*** & 0.254*** & 1.214*** & 1.351*** & -6.183*** & 6.349*** & -5.624*** & -2.819*** & -1.230*** & -0.964*** \\
& (0.0709) & (0.00275) & (0.0281) & (0.0183) & (0.00353) & (0.0362) & (0.0540) & (0.0321) & (0.0231) & (0.0166) & (0.0125) \\
\textbf{Version4} & -47.20*** & -1.456*** & 0.825*** & 3.535*** & 1.429*** & -5.974*** & 8.112*** & -5.505*** & -2.779*** & -1.206*** & -0.960*** \\
& (0.0706) & (0.00274) & (0.0284) & (0.0173) & (0.00351) & (0.0360) & (0.0545) & (0.0320) & (0.0231) & (0.0165) & (0.0125) \\
\textbf{Version5} & -44.54*** & -1.387*** & 0.418*** & 3.235*** & 1.437*** & -5.541*** & 7.907*** & -5.111*** & -2.589*** & -1.127*** & -0.903*** \\
& (0.0696) & (0.00269) & (0.0274) & (0.0166) & (0.00340) & (0.0347) & (0.0526) & (0.0311) & (0.0223) & (0.0160) & (0.0122) \\
\textbf{Version6} & -20.85*** & -0.632*** & -2.377*** & 4.154*** & 0.859*** & -2.620*** & 10.22*** & -4.902*** & -2.590*** & -0.931*** & -0.872*** \\
& (0.0559) & (0.00235) & (0.0273) & (0.0162) & (0.00353) & (0.0320) & (0.0545) & (0.0309) & (0.0224) & (0.0154) & (0.0120) \\
\textbf{Baseline} & 185.9*** & 6.708*** & 25.31*** & 71.40*** & 14.91*** & 28.15*** & 14.60*** & 12.85*** & 4.392*** & 3.881*** & 1.953*** \\
& (0.0557) & (0.00217) & (0.0217) & (0.0131) & (0.00275) & (0.0276) & (0.0411) & (0.0260) & (0.0191) & (0.0133) & (0.0104) \\
\textbf{Article FE} & Yes & Yes & Yes & Yes & Yes & Yes & Yes & Yes & Yes & Yes & Yes \\
\hline
\textbf{R2} & 0.876 & 0.853 & 0.754 & 0.774 & 0.881 & 0.908 & 0.703 & 0.843 & 0.750 & 0.927 & 0.878 \\
\textbf{N} & 4,370,848 & 4,370,848 & 4,370,848 & 4,370,848 & 4,370,848 & 4,370,848 & 4,370,848 & 4,370,848 & 4,370,848 & 4,370,848 & 4,370,848 \\
\hline
\hline
\end{tabular}
}
\end{adjustbox}
\end{table}

\begin{table}[htbp]
\centering
\caption{Writing Rules: Within Paper Comparison with Controls}
\label{tab5_Measure the Difference of original and GPT-revised Texts}
\parbox[t]{5.5in}{\footnotesize{
This table shows the differences in writing rules between the abstracts and prompt specific revised ones following specification \ref{heterogeneity_in_rule_bt_original_revise}. Instead of directly control for the article fixed effect, we include a set of paper-level characteristics: Ethinicitlity , Gender, discipline, Number of Papers, Year of Academic experience (YoE), and date fixed effect. All errors are robust. Coefficients marked with *, **, and *** are significant at the 10\%, 5\%, and 1\% levels, respectively.
}}
\bigskip
\begin{adjustbox}{center, max width=\textwidth}
\scalebox{0.62}{%
\begin{tabular}{lccccccccccc}
\hline\hline
                    &\multicolumn{1}{c}{(1)}&\multicolumn{1}{c}{(2)}&\multicolumn{1}{c}{(3)}&\multicolumn{1}{c}{(4)}&\multicolumn{1}{c}{(5)}&\multicolumn{1}{c}{(6)}&\multicolumn{1}{c}{(7)}&\multicolumn{1}{c}{(8)}&\multicolumn{1}{c}{(9)}&\multicolumn{1}{c}{(10)}&\multicolumn{1}{c}{(11)}\\                    &\multicolumn{1}{c}{Rule1a}&\multicolumn{1}{c}{Rule1b}&\multicolumn{1}{c}{Rule2}&\multicolumn{1}{c}{Rule4}&\multicolumn{1}{c}{Rule5}&\multicolumn{1}{c}{Rule7a}&\multicolumn{1}{c}{Rule7b}&\multicolumn{1}{c}{Rule8}&\multicolumn{1}{c}{Rule9}&\multicolumn{1}{c}{Rule10a}&\multicolumn{1}{c}{Rule10b}\\
\hline
Version1            &      -46.52{***}&      -1.293{***}&       4.050{***}&       4.829{***}&       1.618{***}&      -5.566{***}&       8.503{***}&      -4.953{***}&      -2.608{***}&      -1.080{***}&      -0.921{***}\\
                    &    (0.0690)         &   (0.00268)         &    (0.0276)         &    (0.0161)         &   (0.00348)         &    (0.0348)         &    (0.0540)         &    (0.0314)         &    (0.0227)         &    (0.0164)         &    (0.0125)         \\
Version2            &      -45.14{***}&      -1.349{***}&       1.419{***}&       3.848{***}&       1.435{***}&      -5.330{***}&       5.296{***}&      -5.490{***}&      -2.780{***}&      -1.198{***}&      -0.961{***}\\
                    &    (0.0701)         &   (0.00271)         &    (0.0274)         &    (0.0170)         &   (0.00351)         &    (0.0367)         &    (0.0542)         &    (0.0325)         &    (0.0232)         &    (0.0169)         &    (0.0127)         \\
Version3            &      -47.03{***}&      -1.469{***}&       0.269{***}&       1.222{***}&       1.349{***}&      -6.205{***}&       6.344{***}&      -5.639{***}&      -2.818{***}&      -1.229{***}&      -0.968{***}\\
                    &    (0.0714)         &   (0.00277)         &    (0.0282)         &    (0.0184)         &   (0.00359)         &    (0.0374)         &    (0.0543)         &    (0.0325)         &    (0.0232)         &    (0.0170)         &    (0.0127)         \\
Version4            &      -47.28{***}&      -1.459{***}&       0.841{***}&       3.534{***}&       1.427{***}&      -5.995{***}&       8.111{***}&      -5.521{***}&      -2.778{***}&      -1.206{***}&      -0.964{***}\\
                    &    (0.0711)         &   (0.00276)         &    (0.0285)         &    (0.0174)         &   (0.00357)         &    (0.0373)         &    (0.0547)         &    (0.0324)         &    (0.0231)         &    (0.0171)         &    (0.0127)         \\
Version5            &      -44.62{***}&      -1.389{***}&       0.432{***}&       3.235{***}&       1.435{***}&      -5.562{***}&       7.905{***}&      -5.124{***}&      -2.588{***}&      -1.127{***}&      -0.906{***}\\
                    &    (0.0701)         &   (0.00271)         &    (0.0275)         &    (0.0167)         &   (0.00346)         &    (0.0360)         &    (0.0529)         &    (0.0315)         &    (0.0224)         &    (0.0165)         &    (0.0124)         \\
Version6            &      -20.93{***}&      -0.635{***}&      -2.364{***}&       4.153{***}&       0.857{***}&      -2.628{***}&       10.21{***}&      -4.918{***}&      -2.587{***}&      -0.930{***}&      -0.876{***}\\
                    &    (0.0563)         &   (0.00236)         &    (0.0274)         &    (0.0163)         &   (0.00357)         &    (0.0334)         &    (0.0548)         &    (0.0313)         &    (0.0224)         &    (0.0160)         &    (0.0122)         \\
Africans            &     -0.0628         &      0.0274         &       0.631{**} &       0.288         &      -0.149{***}&      -1.592{***}&       0.100         &      0.0219         &    -0.00632         &      -0.303         &      -0.291{**} \\
                    &     (0.705)         &    (0.0239)         &     (0.232)         &     (0.153)         &    (0.0443)         &     (0.465)         &     (0.375)         &     (0.297)         &     (0.156)         &     (0.209)         &    (0.0989)         \\
British             &       5.153{***}&       0.125{***}&      -0.692{***}&       0.186{*}  &     0.00529         &       0.296         &       1.183{***}&       0.705{***}&       0.388{***}&     -0.0691         &     -0.0567         \\
                    &     (0.343)         &    (0.0115)         &     (0.109)         &    (0.0729)         &    (0.0208)         &     (0.223)         &     (0.178)         &     (0.142)         &    (0.0768)         &     (0.101)         &    (0.0535)         \\
EastAsian           &     -0.0569         &       0.194{***}&       2.127{***}&       1.627{***}&       0.775{***}&       3.205{***}&       3.043{***}&      -1.586{***}&      -0.165{*}  &      -0.756{***}&     -0.0341         \\
                    &     (0.308)         &    (0.0103)         &    (0.0969)         &    (0.0668)         &    (0.0188)         &     (0.201)         &     (0.166)         &     (0.124)         &    (0.0666)         &    (0.0940)         &    (0.0498)         \\
EastEuropean        &      -7.721{***}&      -0.233{***}&       1.173{***}&      -0.219{*}  &      0.0719{*}  &      -1.278{***}&      -2.483{***}&      -0.230         &      0.0538         &      -0.624{***}&      -0.193{**} \\
                    &     (0.483)         &    (0.0159)         &     (0.156)         &     (0.105)         &    (0.0298)         &     (0.305)         &     (0.235)         &     (0.194)         &     (0.108)         &     (0.134)         &    (0.0716)         \\
Indian              &       4.479{***}&       0.136{***}&       0.265{*}  &       0.436{***}&      -0.169{***}&      -3.426{***}&      -0.102         &       0.123         &      -0.275{***}&      0.0495         &      0.0512         \\
                    &     (0.396)         &    (0.0133)         &     (0.124)         &    (0.0837)         &    (0.0236)         &     (0.252)         &     (0.211)         &     (0.163)         &    (0.0810)         &     (0.127)         &    (0.0665)         \\
Jewish              &       0.358         &      0.0468{**} &       0.786{***}&       0.203         &      0.0147         &      -0.773{*}  &      -0.189         &       0.860{***}&       0.356{**} &      -0.440{**} &      -0.178{*}  \\
                    &     (0.541)         &    (0.0182)         &     (0.174)         &     (0.115)         &    (0.0328)         &     (0.349)         &     (0.273)         &     (0.226)         &     (0.123)         &     (0.155)         &    (0.0818)         \\
Muslim              &       4.032{***}&       0.152{***}&       0.570{***}&      0.0356         &      -0.296{***}&      -1.205{***}&       2.644{***}&     -0.0272         &      -0.183         &      -0.968{***}&     -0.0872         \\
                    &     (0.491)         &    (0.0169)         &     (0.161)         &     (0.107)         &    (0.0305)         &     (0.335)         &     (0.276)         &     (0.208)         &     (0.104)         &     (0.143)         &    (0.0774)         \\
Biology    &       18.44{***}&       1.540{***}&       5.899{***}&      -2.265{***}&       0.879{***}&       6.212{***}&       20.76{***}&     -0.0878         &      -0.866{***}&       2.498{***}&       0.232{**} \\
                    &     (0.656)         &    (0.0255)         &     (0.229)         &     (0.179)         &    (0.0431)         &     (0.478)         &     (0.437)         &     (0.278)         &     (0.142)         &     (0.225)         &    (0.0800)         \\
Computer Science&       21.93{***}&       1.725{***}&       5.877{***}&       1.683{***}&      0.0889{***}&       14.78{***}&       19.27{***}&       1.612{***}&      -1.119{***}&       2.137{***}&       0.885{***}\\
                    &     (0.217)         &   (0.00721)         &    (0.0712)         &    (0.0460)         &    (0.0136)         &     (0.143)         &     (0.102)         &    (0.0917)         &    (0.0515)         &    (0.0495)         &    (0.0273)         \\
Economics  &       6.332{***}&       0.992{***}&       5.455{***}&       3.120{***}&      -0.708{***}&       4.613{***}&       11.36{***}&       0.509         &      -0.622{***}&       0.408{**} &     -0.0209         \\
                    &     (0.709)         &    (0.0275)         &     (0.283)         &     (0.185)         &    (0.0515)         &     (0.585)         &     (0.496)         &     (0.361)         &     (0.185)         &     (0.151)         &    (0.0667)         \\
Electrical Eng. &       23.85{***}&       1.799{***}&       5.104{***}&      -2.250{***}&       0.171{***}&       10.91{***}&       18.72{***}&      0.0548         &      -1.549{***}&       3.367{***}&       1.726{***}\\
                    &     (0.335)         &    (0.0127)         &     (0.124)         &    (0.0781)         &    (0.0233)         &     (0.277)         &     (0.226)         &     (0.157)         &    (0.0709)         &     (0.138)         &    (0.0836)         \\
Finance    &       7.212{***}&       1.009{***}&       4.774{***}&       1.293{***}&      -1.052{***}&       5.788{***}&       12.55{***}&      -0.103         &      -0.760{***}&       0.941{***}&       0.563{***}\\
                    &     (0.890)         &    (0.0337)         &     (0.346)         &     (0.211)         &    (0.0597)         &     (0.727)         &     (0.618)         &     (0.418)         &     (0.223)         &     (0.243)         &     (0.158)         \\
Physics    &       31.77{***}&       1.440{***}&      -1.056{***}&      -2.548{***}&       0.611{***}&       2.356{***}&       15.10{***}&      -0.461{***}&      -0.662{***}&       3.662{***}&       1.110{***}\\
                    &     (0.225)         &   (0.00708)         &    (0.0685)         &    (0.0470)         &    (0.0134)         &     (0.131)         &    (0.0976)         &    (0.0872)         &    (0.0524)         &    (0.0560)         &    (0.0299)         \\
Statistics &       21.03{***}&       1.673{***}&       5.981{***}&      -0.260{**} &       0.423{***}&       13.87{***}&       11.33{***}&       1.250{***}&      -1.463{***}&       1.776{***}&       0.140{**} \\
                    &     (0.406)         &    (0.0154)         &     (0.148)         &    (0.0922)         &    (0.0275)         &     (0.347)         &     (0.253)         &     (0.202)         &    (0.0838)         &     (0.130)         &    (0.0479)         \\
Female              &       0.597         &     -0.0244         &      -1.529{***}&       0.385{***}&      -0.209{***}&      -3.212{***}&       0.535{*}  &       0.542{**} &       0.463{***}&       0.470{***}&     -0.0145         \\
                    &     (0.419)         &    (0.0142)         &     (0.131)         &    (0.0916)         &    (0.0253)         &     (0.270)         &     (0.224)         &     (0.173)         &    (0.0949)         &     (0.129)         &    (0.0670)         \\
Male                &      -7.949{***}&      -0.205{***}&      0.0322         &       1.423{***}&       0.145{***}&      -1.233{***}&      -2.416{***}&      -0.500{***}&      0.0445         &      -0.186{*}  &      -0.126{**} \\
                    &     (0.254)         &   (0.00847)         &    (0.0780)         &    (0.0546)         &    (0.0151)         &     (0.160)         &     (0.134)         &     (0.101)         &    (0.0546)         &    (0.0779)         &    (0.0415)         \\
No. of Papers&       3.692{***}&      0.0828{***}&      -0.306{***}&   -0.000276         &    -0.00893         &       0.612{***}&       0.820{***}&       0.268{***}&     0.00977         &       0.101{***}&       0.144{***}\\
                    &    (0.0970)         &   (0.00318)         &    (0.0288)         &    (0.0199)         &   (0.00589)         &    (0.0629)         &    (0.0541)         &    (0.0383)         &    (0.0194)         &    (0.0292)         &    (0.0192)         \\
YoE     &      -1.131{***}&      -0.110{***}&      -0.476{***}&       0.226{***}&      0.0190{**} &      -0.994{***}&      -2.188{***}&      0.0809{*}  &       0.193{***}&      -0.177{***}&    -0.00452         \\
                    &     (0.103)         &   (0.00337)         &    (0.0314)         &    (0.0217)         &   (0.00611)         &    (0.0620)         &    (0.0505)         &    (0.0390)         &    (0.0215)         &    (0.0280)         &    (0.0160)         \\
Baseline            &       167.6{***}&       5.449{***}&       22.56{***}&       70.54{***}&       14.40{***}&       21.68{***}&       1.002{***}&       12.92{***}&       5.078{***}&       1.980{***}&       1.273{***}\\
                    &     (0.348)         &    (0.0114)         &     (0.107)         &    (0.0728)         &    (0.0205)         &     (0.211)         &     (0.166)         &     (0.137)         &    (0.0790)         &    (0.0915)         &    (0.0495)         \\
\hline
R2                  &       0.113         &       0.144         &      0.0371         &      0.0332         &      0.0315         &      0.0325         &      0.0465         &     0.00665         &     0.00613         &     0.00548         &     0.00339         \\
N                   &     4379544         &     4379544         &     4379544         &     4379544         &     4379544         &     4379544         &     4379544         &     4379544         &     4379544         &     4379544         &     4379544         \\
\hline\hline
\textbf{Date FE} & Yes & Yes & Yes & Yes & Yes & Yes & Yes & Yes & Yes & Yes & Yes \\
\textbf{Obs.} & 4,379,544 & 4,379,544 & 4,379,544 & 4,379,544 & 4,379,544 & 4,379,544 & 4,379,544 & 4,379,544 & 4,379,544 & 4,379,544 & 4,379,544 \\
\textbf{R-squared} & 0.110 & 0.142 & 0.037 & 0.033 & 0.032 & 0.032 & 0.046 & 0.007 & 0.006 & 0.004 & 0.003 \\
\hline
\hline
\end{tabular}
}
\end{adjustbox}
\end{table}

\begin{table}[htbp]
\centering
\caption{Writing Rules between of Texts with and without Revisions}
\label{tab7_Difference in the Writing Rules between Human and GPT-written}
\parbox[t]{5.5in}{\footnotesize{
This table examines the differences in writing rules between unrevised and revised texts, where the classification of the texts is based on our multi-label classification model. We report the regression results following specification \ref{heterogeneity_of_rule_bt_unrevised_revised}. Since each abstract is treated as a single observation, we control for article-level characteristics rather than using article-fixed effects. All errors are robust. Coefficients marked with *, **, and *** are significant at 10\%, 5\%, and 1\%, respectively.
}}
\bigskip
\begin{adjustbox}{center, max width=\textwidth}
\scalebox{0.62}{%
\begin{tabular}{lccccccccccc}
\hline\hline
                    &\multicolumn{1}{c}{(1)}&\multicolumn{1}{c}{(2)}&\multicolumn{1}{c}{(3)}&\multicolumn{1}{c}{(4)}&\multicolumn{1}{c}{(5)}&\multicolumn{1}{c}{(6)}&\multicolumn{1}{c}{(7)}&\multicolumn{1}{c}{(8)}&\multicolumn{1}{c}{(9)}&\multicolumn{1}{c}{(10)}&\multicolumn{1}{c}{(11)}\\
                    &\multicolumn{1}{c}{Rule1a}&\multicolumn{1}{c}{Rule1b}&\multicolumn{1}{c}{Rule2}&\multicolumn{1}{c}{Rule4}&\multicolumn{1}{c}{Rule5}&\multicolumn{1}{c}{Rule7a}&\multicolumn{1}{c}{Rule7b}&\multicolumn{1}{c}{Rule8}&\multicolumn{1}{c}{Rule9}&\multicolumn{1}{c}{Rule10a}&\multicolumn{1}{c}{Rule10b}\\
\hline
Prediction 1        & -26.73{***}  & -0.610{***}  &  3.623{***}  &  2.583{***}  & -0.126{***}  & -1.851{***}  &  3.817{***}  & -3.144{***}  & -1.700{***}  & -0.616{***}  & -0.858{***} \\
                    & (0.386)          & (0.0138)         & (0.128)          & (0.0838)         & (0.0220)         & (0.250)          & (0.219)          & (0.166)          & (0.0974)         & (0.123)          & (0.0663)         \\
Prediction 2        & -12.81{***}  & -0.150{***}  &  0.607{*}    &  2.918{***}  &  0.0943{*}   &  5.384{***}  & 10.40{***}   & -4.117{***}  & -1.704{***}  & -1.148{***}  & -1.159{***} \\
                    & (0.754)          & (0.0288)         & (0.242)          & (0.163)          & (0.0433)         & (0.599)          & (0.553)          & (0.361)          & (0.201)          & (0.250)          & (0.139)         \\
Prediction 3        & -43.89{***}  & -1.346{***}  &  1.258{**}   & -6.433{***}  & -0.400{***}  & -1.197           &  1.535{*}    & -5.157{***}  & -2.678{***}  & -1.661{***}  & -1.045{***} \\
                    & (1.293)          & (0.0463)         & (0.448)          & (0.284)          & (0.0791)         & (0.804)          & (0.663)          & (0.474)          & (0.267)          & (0.325)          & (0.176)         \\
Prediction 4        & -37.55{***}  & -0.968{***}  &  1.463{**}   &  2.775{***}  & -0.344{***}  &  1.490           &  8.328{***}  & -6.255{***}  & -2.131{***}  & -0.860           & -1.451{***} \\
                    & (1.266)          & (0.0500)         & (0.451)          & (0.297)          & (0.0773)         & (0.995)          & (0.905)          & (0.564)          & (0.310)          & (0.442)          & (0.207)         \\
Prediction 5        & -39.54{***}  & -1.323{***}  & -1.647{***}  &  1.617{***}  &  0.175{***}  & -1.296{***}  &  2.304{***}  & -3.573{***}  & -1.174{***}  & -0.805{***}  & -0.732{***} \\
                    & (0.458)          & (0.0170)         & (0.173)          & (0.120)          & (0.0310)         & (0.370)          & (0.309)          & (0.240)          & (0.147)          & (0.163)          & (0.101)         \\
Prediction 6        &  1.944{***}  &  0.133{***}  & -3.782{***}  &  3.116{***}  &  0.250{***}  &  8.790{***}  & 17.03{***}   & -3.905{***}  & -1.665{***}  & -0.593{*}    & -1.002{***} \\
                    & (0.544)          & (0.0221)         & (0.186)          & (0.132)          & (0.0349)         & (0.538)          & (0.519)          & (0.322)          & (0.174)          & (0.241)          & (0.135)         \\
Africans             & -1.029           &  0.0822{**}  &  1.883{***}  & -0.104           & -0.249{***}  & -1.687{**}   &  0.335           & -0.0387          &  0.200           & -0.151           & -0.443{**}  \\
                    & (0.869)          & (0.0300)         & (0.270)          & (0.179)          & (0.0464)         & (0.519)          & (0.391)          & (0.380)          & (0.250)          & (0.261)          & (0.145)         \\
British              &  5.923{***}  &  0.197{***}  & -0.413{***}  &  1.142{***}  &  0.0452{*}   &  0.577{*}    &  1.613{***}  &  0.494{**}   &  0.568{***}  & -0.0558          & -0.159{*}   \\
                    & (0.419)          & (0.0144)         & (0.125)          & (0.0837)         & (0.0217)         & (0.248)          & (0.183)          & (0.180)          & (0.123)          & (0.121)          & (0.0728)        \\
EastAsian            & -0.837{*}    &  0.182{***}  &  2.016{***}  &  1.174{***}  &  0.874{***}  &  3.854{***}  &  1.291{***}  & -1.424{***}  &  0.0656          & -0.771{***}  &  0.105          \\
                    & (0.372)          & (0.0129)         & (0.111)          & (0.0761)         & (0.0197)         & (0.223)          & (0.170)          & (0.159)          & (0.110)          & (0.111)          & (0.0672)        \\
EastEuropean         & -12.87{***}  & -0.235{***}  &  3.644{***}  & -1.348{***}  &  0.0893{**}  & -1.907{***}  & -1.776{***}  & -0.579{*}    &  0.0857          & -0.666{***}  & -0.309{**}  \\
                    & (0.595)          & (0.0204)         & (0.188)          & (0.123)          & (0.0314)         & (0.337)          & (0.237)          & (0.246)          & (0.172)          & (0.162)          & (0.0961)        \\
Indian               &  4.080{***}  &  0.236{***}  &  1.675{***}  &  0.404{***}  & -0.270{***}  & -4.356{***}  &  1.049{***}  & -0.519{*}    & -0.507{***}  &  0.0933          & -0.0552         \\
                    & (0.482)          & (0.0167)         & (0.144)          & (0.0960)         & (0.0246)         & (0.280)          & (0.221)          & (0.207)          & (0.133)          & (0.149)          & (0.0885)        \\
Jewish               & -0.622           &  0.110{***}  &  1.788{***}  &  0.499{***}  &  0.106{**}   & -1.286{***}  &  0.493           &  0.474           &  0.481{*}    & -0.447{*}    & -0.0866         \\
                    & (0.665)          & (0.0229)         & (0.203)          & (0.134)          & (0.0343)         & (0.385)          & (0.282)          & (0.284)          & (0.193)          & (0.188)          & (0.116)         \\
Muslim               &  7.182{***}  &  0.398{***}  &  2.480{***}  & -0.952{***}  & -0.512{***}  & -1.553{***}  &  2.483{***}  & -0.0597          & -0.180           & -1.034{***}  & -0.247{*}   \\
                    & (0.605)          & (0.0214)         & (0.186)          & (0.125)          & (0.0316)         & (0.372)          & (0.293)          & (0.270)          & (0.173)          & (0.177)          & (0.109)         \\
Biology     & 58.84{***}   &  2.868{***}  &  1.242{***}  & -5.539{***}  &  1.199{***}  & 10.81{***}   & 13.87{***}   &  4.451{***}  &  0.807{**}   &  3.720{***}  &  0.750{***} \\
                    & (0.789)          & (0.0306)         & (0.241)          & (0.195)          & (0.0428)         & (0.565)          & (0.490)          & (0.405)          & (0.269)          & (0.304)          & (0.136)         \\
Computer Science & 49.57{***}  &  2.669{***}  &  2.395{***}  & -0.140{**}   &  0.446{***}  & 19.33{***}   & 13.47{***}   &  6.335{***}  & -0.0312          &  3.008{***}  &  2.118{***} \\
                    & (0.265)          & (0.00899)        & (0.0833)         & (0.0543)         & (0.0143)         & (0.158)          & (0.109)          & (0.115)          & (0.0749)         & (0.0644)         & (0.0434)        \\
Economics   & 19.35{***}   &  1.529{***}  &  4.738{***}  &  2.413{***}  &  0.0299          &  6.669{***}   &  9.422{***}   &  2.264{***}  &  0.251           &  0.837{***}  &  0.175          \\
                    & (0.922)          & (0.0351)         & (0.326)          & (0.216)          & (0.0526)         & (0.651)          & (0.535)          & (0.455)          & (0.302)          & (0.235)          & (0.121)         \\
Electrical Eng. & 43.85{***}  &  2.447{***}  &  1.359{***}  & -6.048{***}  &  0.334{***}  & 13.19{***}   & 11.25{***}   &  3.111{***}  & -1.091{***}  &  4.096{***}  &  2.518{***} \\
                    & (0.407)          & (0.0154)         & (0.136)          & (0.0915)         & (0.0236)         & (0.307)          & (0.239)          & (0.208)          & (0.118)          & (0.163)          & (0.107)         \\
Finance     & 23.24{***}   &  1.547{***}  &  2.114{***}  & -0.714{**}   & -0.554{***}  &  7.339{***}  &  9.070{***}  &  2.587{***}  & -0.0106          &  1.385{***}  &  1.112{***} \\
                    & (1.180)          & (0.0442)         & (0.396)          & (0.254)          & (0.0621)         & (0.811)          & (0.651)          & (0.564)          & (0.366)          & (0.318)          & (0.231)         \\
Physics     & 47.37{***}   &  1.914{***}  & -3.146{***}  & -4.779{***}  &  0.729{***}  &  3.854{***}  &  9.457{***}  &  0.998{***}  &  0.702{***}  &  3.985{***}  &  1.345{***} \\
                    & (0.272)          & (0.00879)        & (0.0812)         & (0.0545)         & (0.0142)         & (0.144)          & (0.0984)         & (0.105)          & (0.0757)         & (0.0658)         & (0.0377)        \\
Statistics  & 42.09{***}   &  2.394{***}  &  2.817{***}  & -2.270{***}  &  0.899{***}  & 16.58{***}   &  8.396{***}   &  3.832{***}  & -0.675{***}  &  2.762{***}  &  0.561{***} \\
                    & (0.486)          & (0.0184)         & (0.164)          & (0.106)          & (0.0279)         & (0.380)          & (0.271)          & (0.260)          & (0.153)          & (0.172)          & (0.0836)        \\
Female               &  2.261{***}  & -0.0217          & -2.211{***}  &  1.541{***}  & -0.139{***}  & -2.973{***}  &  0.471{*}    &  0.284           &  0.871{***}  &  0.581{***}  & -0.0288         \\
                    & (0.511)          & (0.0176)         & (0.151)          & (0.104)          & (0.0264)         & (0.302)          & (0.234)          & (0.220)          & (0.151)          & (0.154)          & (0.0915)        \\
Male                 & -11.31{***}  & -0.332{***}  & -0.286{**}   &  2.508{***}  &  0.365{***}  & -1.708{***}  & -1.803{***}  & -1.054{***}  &  0.256{**}   & -0.102           & -0.103          \\
                    & (0.307)          & (0.0106)         & (0.0902)         & (0.0620)         & (0.0159)         & (0.178)          & (0.138)          & (0.128)          & (0.0891)         & (0.0914)         & (0.0543)        \\
No. of Papers &  4.131{***}  &  0.0735{***}  & -0.528{***}  &  0.0847{***}  &  0.00219         &  0.756{***}  &  0.814{***}  &  0.337{***}  & -0.0606          &  0.129{***}  &  0.180{***} \\
                    & (0.113)          & (0.00379)        & (0.0326)         & (0.0227)         & (0.00618)        & (0.0700)         & (0.0564)         & (0.0495)         & (0.0325)         & (0.0348)         & (0.0246)        \\
YoE      & -3.126{***}  & -0.182{***}  & -0.0214          &  0.804{***}  &  0.0246{***}  & -1.261{***}  & -1.476{***}  & -0.175{***}  &  0.391{***}  & -0.312{***}  & -0.0175         \\
                    & (0.125)          & (0.00420)        & (0.0371)         & (0.0251)         & (0.00647)        & (0.0687)         & (0.0513)         & (0.0495)         & (0.0352)         & (0.0336)         & (0.0209)        \\
Baseline             & 154.8{***}   &  4.976{***}  & 24.46{***}   & 71.47{***}   & 14.07{***}   & 19.41{***}   &  4.883{***}   & 11.25{***}   &  4.087{***}  &  1.513{***}  &  0.793{***} \\
                    & (0.416)          & (0.0140)         & (0.123)          & (0.0834)         & (0.0215)         & (0.231)          & (0.166)          & (0.168)          & (0.117)          & (0.108)          & (0.0647)        \\
\hline
R3                 &  0.107           &  0.175           &  0.0216          &  0.0410          &  0.0166          &  0.0442          &  0.0345          &  0.00836         &  0.00208         &  0.00489         &  0.00424        \\
N                    &  627225          &  627225          &  627225          &  627225          &  627225          &  627225          &  627225          &  627225          &  627225          &  627225          &  627225         \\
\hline\hline
\textbf{Article FE}  & No               & No               & No               & No               & No               & No               & No               & No               & No               & No               & No              \\
\textbf{Time FE}     & Yes              & Yes              & Yes              & Yes              & Yes              & Yes              & Yes              & Yes              & Yes              & Yes              & Yes             \\
\textbf{Obs.}        & 627,225          & 627,225          & 627,225          & 627,225          & 627,225          & 627,225          & 627,225          & 627,225          & 627,225          & 627,225          & 627,225         \\
\textbf{R-squared}   & 0.099            & 0.168            & 0.021            & 0.040            & 0.017            & 0.044            & 0.034            & 0.008            & 0.002            & 0.004            & 0.003           \\
\hline
\hline
\end{tabular}
}
\end{adjustbox}
\end{table}

\clearpage

\clearpage

\clearpage

\clearpage

\clearpage

\begin{figure}[H]
\begin{center}
   \caption{Heterogeneity in difference-in-difference of Textual Similarity (Seniority)}  
   \vspace{-2pt}
   \label{figure11_similarity_bt_junior_senior}
   
   \begin{subfigure}[t]{0.49\textwidth}
       \centering
       \includegraphics[width=\textwidth, height=0.4\textheight]{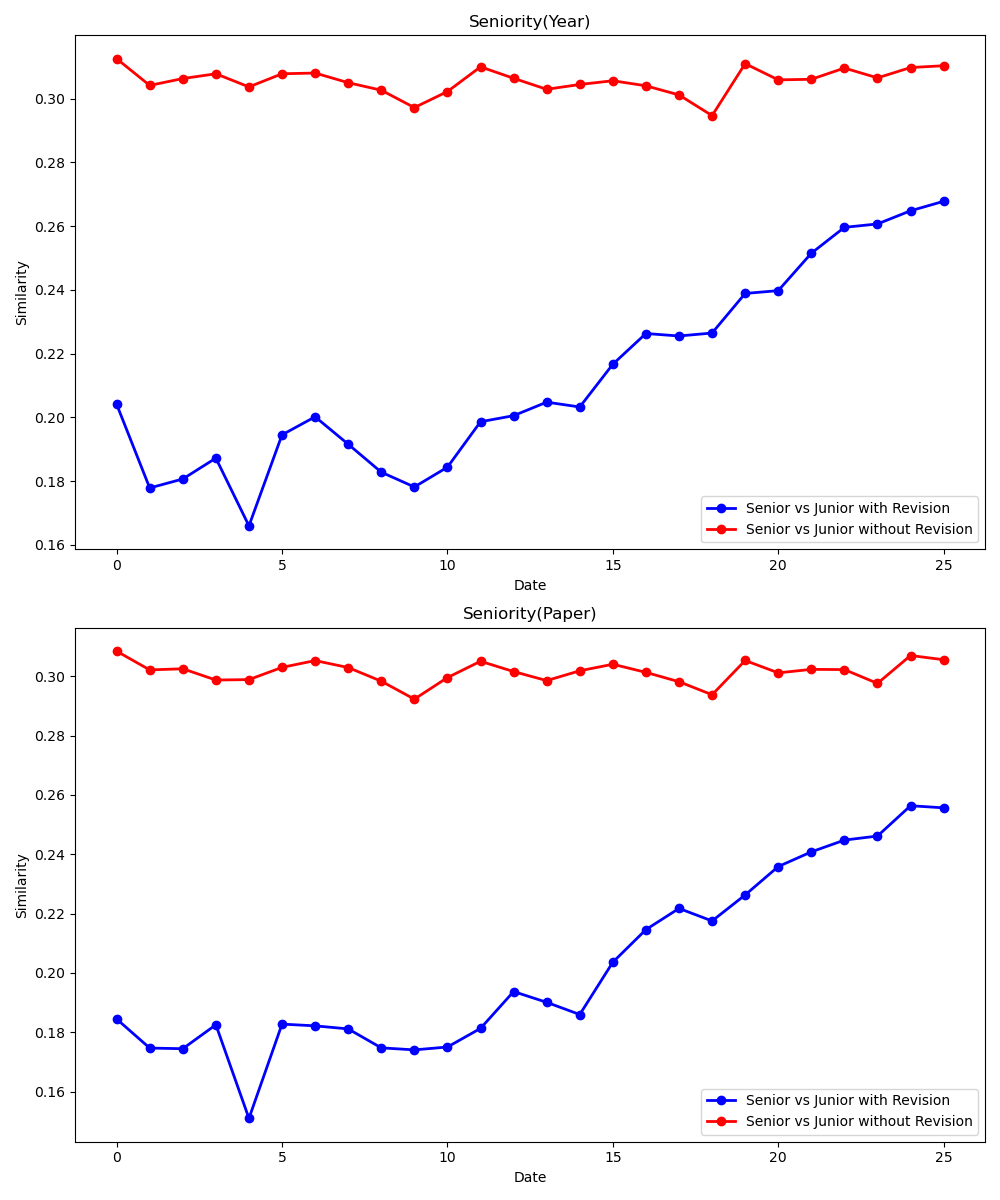}
       \caption*{Left Panel}
   \end{subfigure}
   \hfill
   \begin{subfigure}[t]{0.49\textwidth}
       \centering
       \includegraphics[width=\textwidth, height=0.4\textheight]{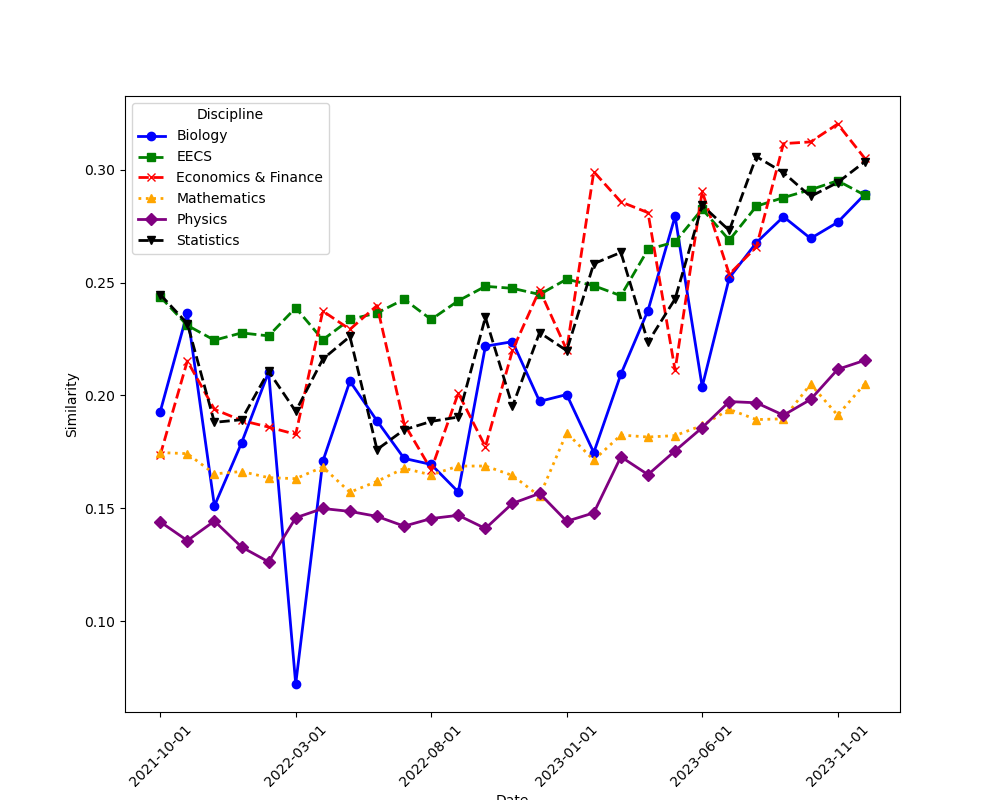}
       \caption*{Right Panel}
   \end{subfigure}
   \vspace{-2pt}
\end{center}   

{\footnotesize 
Note. This figure presents a difference-in-difference analysis of textual similarity between senior and junior authors. The Left Panel depicts the overall textual similarity between senior and junior authors. The red line represents similarity between the senior and junior authors without GPT adoption, while the blue line shows similarity between the senior and junior authors with GPT adoption. The top subplot uses academic years to define seniority, while the bottom subplot uses the number of published papers as the measure. The Right Panel displays the similarity between senior and junior authors with GPT adoption, broken down by discipline. To compute similarity, a bag-of-words representation is constructed for each article, averaging the representations of abstracts written by junior and senior authors separately within each month and discipline. Authors are classified as senior if they published academic papers prior to 2011, with at least 10 years of research experience; otherwise, they are classified as junior. A paper is considered senior-authored if at least one author is senior.
}   
\end{figure}
\clearpage

\begin{figure}[H]
\begin{center}
   \caption{Heterogeneity in difference-in-difference of Textual Similarity}  
   \vspace{-2pt}
   \label{figure11_similarity_bt_junior_senior}
   \begin{subfigure}[t]{0.49\textwidth}
       \centering
       \includegraphics[width=\textwidth, height=0.4\textheight]{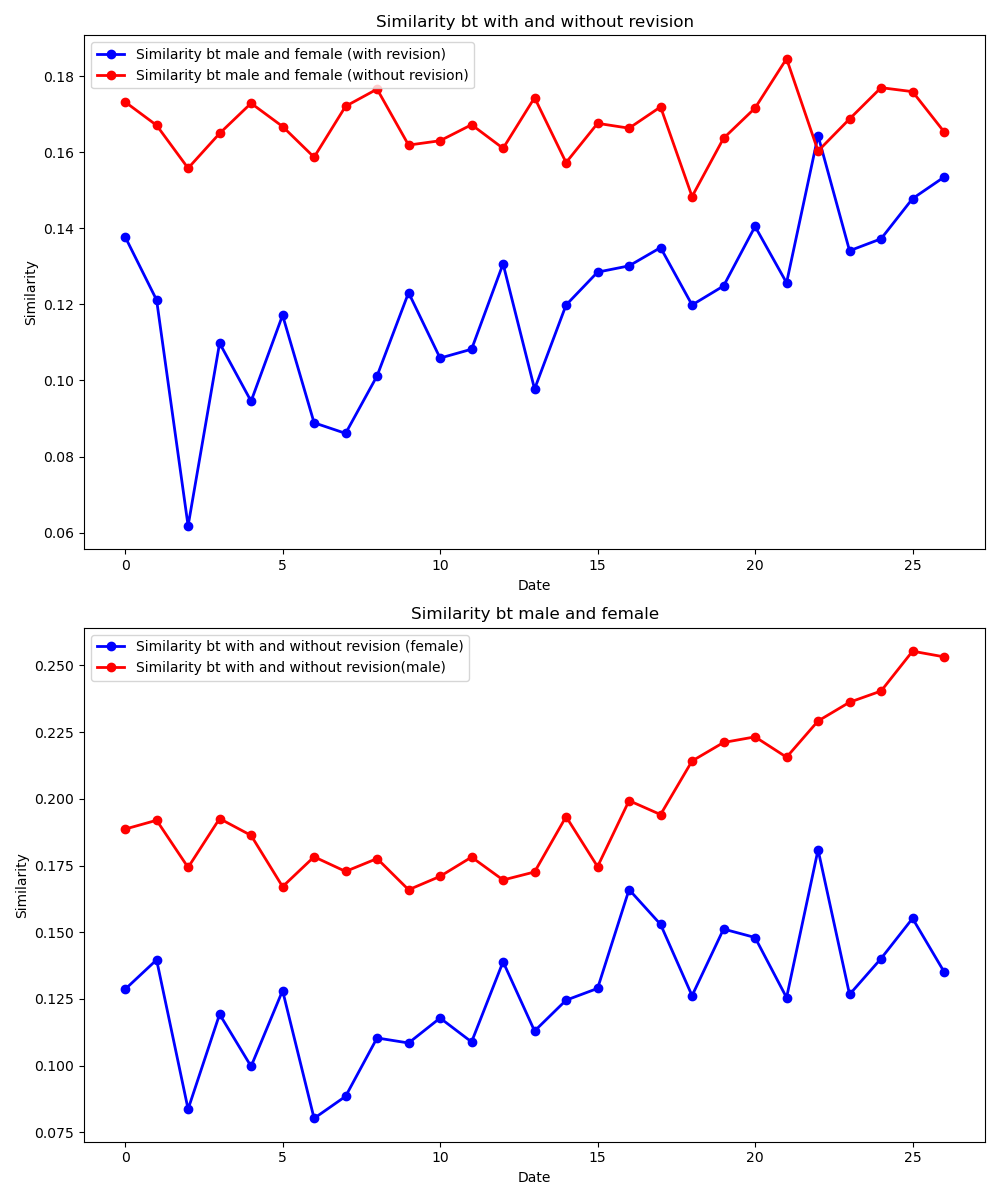}
       \caption*{Left Panel: Gender-based Analysis}
   \end{subfigure}
   \begin{subfigure}[t]{0.49\textwidth}
       \centering
       \includegraphics[width=\textwidth, height=0.4\textheight]{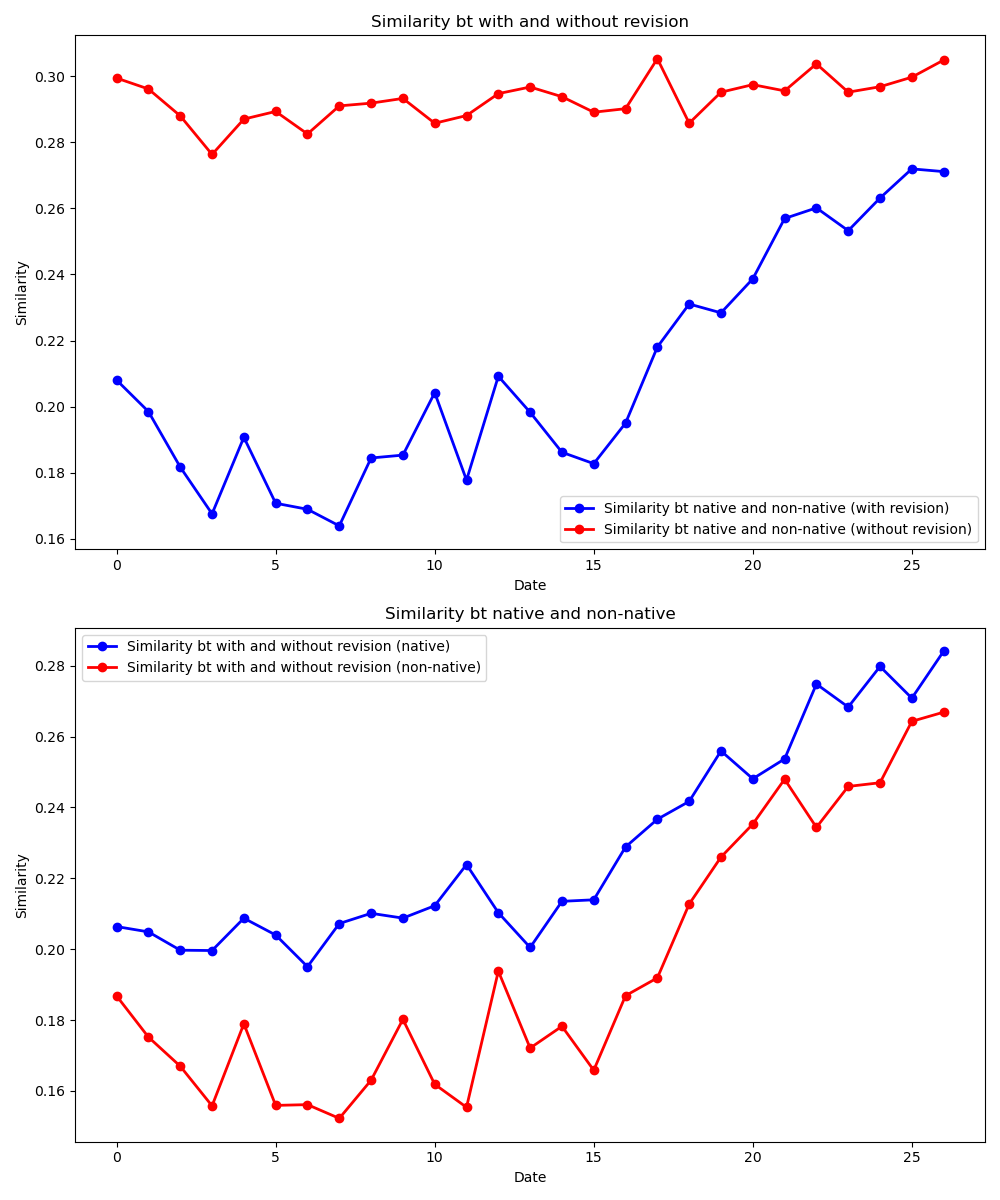}
       \caption*{Right Panel: Native-based Analysis}
   \end{subfigure}
   \vspace{-2pt}
\end{center}   

{\footnotesize 
Note. This figure presents a difference-in-difference analysis of textual similarity. The Left Panel shows the similarity between male and female authors, while the Right Panel contrasts native and non-native authors. For the Left Panel, the top subplot shows the similarity between male and female. The red solid line shows the similarity between male and female without GPT-adoption while the blue line shows that with GPT-adoption. The bottom subplot shows the similarity between those with GPT-adoption and without GPT-adoption conditional on the gender type, the red line despicts the male while the blue line shows the female. For the Right Panel, the top subplot present the similarity between native and non-native conditional on GPT-adoption, the red line shows the results for the cohort without GPT-adoption while the blue line shows the cohort with GPT-adoption. The bottom shows the similarity between those adopting and not adopting GPT conditional on the gender. To compute similarity, we first construct a bag-of-words representation for each article and then average the representations of abstracts within each month and discipline. Authors are classified by gender or language background, and the figure illustrates changes in writing style over time.
}   
\end{figure}
\clearpage

\appendix
\renewcommand{\thefigure}{A\arabic{figure}} 
\setcounter{figure}{0} 

\begin{figure}[H]
\begin{center}
\caption{Boxplot for the Classification Accuracy Distribution}  
\vspace{-2pt}
\label{figure_classification_accuracy}
\includegraphics[width=\textwidth, height=0.6\textheight]{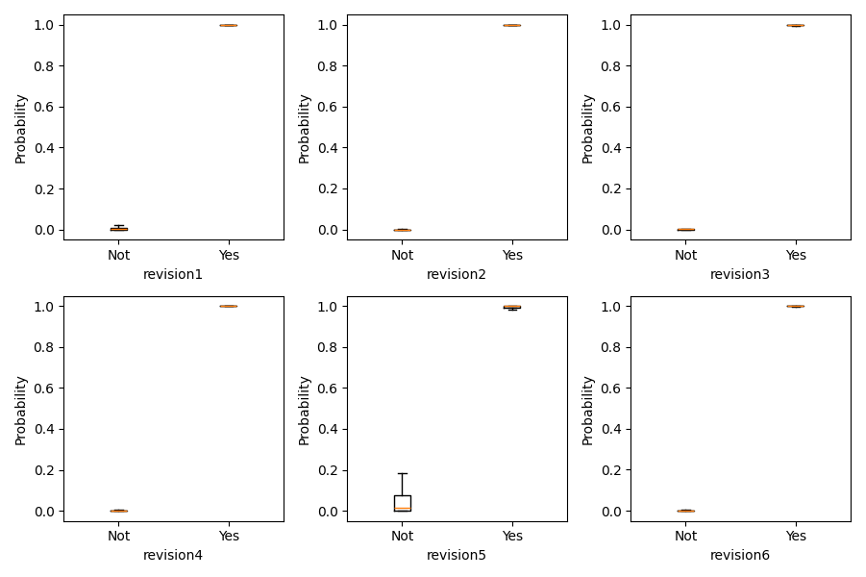}
\vspace{-2pt}
\end{center}
{\footnotesize 
\textit{Note}: Using the text sample, this figure shows the distribution of the estimated probabilities \(P(\hat{y}_i = 1 \mid y_i = 0)\), which is the estimated probability of GPT-revised conditional on human-written, and \(P(\hat{y}_i = 1 \mid y_i = 1)\), the estimated probability of GPT-revised conditional on GPT-revised. The distributions are well-separated, concentrated, and far from 0.5 for each revised version, indicating that our trained models can predict each article with high accuracy.
}
\end{figure}
\clearpage

\begin{table}[htbp]\centering

\caption{Summary Statistics}
\label{tab_app_summary_statistics}
\parbox[t]{5.5in}{\footnotesize{ This table shows the summary statistics for the writing rules and control variables of the abstracts posted on arVix between Jan 1,2021 and Dec 31,2023.
}
}
\begin{tabular}{l*{1}{cccccc}}
\hline\hline
                    \multicolumn{7}{c}{Panel A: Summary Statistics of the Writing Rules}                                                                \\
                    \hline \hline
                    &        Mean&          SD&         P25&         P75&         Min&         Max\\
\hline
Rule1a              &    185.935 &    71.102  &        136 &        232 &         29 &        364 \\
Rule1b              &      6.712 &     2.610  &          5 &          8 &          1 &         13 \\
Rule2               &     25.308 &    20.525  &         10 &         40 &          0 &        100 \\
Rule4               &     71.399 &    14.014  &      62.500 &      80.769 &      28.571 &     100 \\
Rule5               &     14.915 &     3.632  &      12.432 &      17.277 &       6.452 &      27.193 \\
Rule7a              &     28.186 &    44.991  &       0.000 &     100.000 &       0.000 &     100 \\
Rule7b              &     14.615 &    35.326  &       0.000 &       0.000 &       0.000 &     100 \\
Rule8               &     12.866 &    31.277  &       0.000 &       0.000 &       0.000 &     100 \\
Rule9               &      4.390 &    20.487  &       0.000 &       0.000 &       0.000 &     100 \\
Rule10a             &      3.881 &    22.496  &       0.000 &       0.000 &       0.000 &     200 \\
Rule10b             &      1.956 &    13.849  &       0.000 &       0.000 &       0.000 &     100 \\
\hline 
\hline
                    \multicolumn{7}{c}{Summary Staitistics of the Writing Rules}  \\
                    \hline
                    \hline
Africans            &      0.027 &     0.106  &          0 &          0 &          0 &          1 \\
British             &      0.173 &     0.261  &          0 &      0.333 &          0 &          1 \\
East Asian          &      0.293 &     0.398  &          0 &        0.6 &          0 &          1 \\
East European       &      0.056 &     0.167  &          0 &          0 &          0 &          1 \\
Indian              &      0.108 &     0.228  &          0 &        0.1 &          0 &          1 \\
Jewish              &      0.052 &     0.147  &          0 &          0 &          0 &          1 \\
Muslim              &      0.053 &     0.158  &          0 &          0 &          0 &          1 \\
West European       &     .238737&    .3183691&          0  &          0.44  &           0&           1\\
Biology             &      0.011 &     0.104  &          0 &          0 &          0 &          1 \\
Computer Science    &      0.349 &     0.477  &          0 &          1 &          0 &          1 \\
Economics           &      0.007 &     0.086  &          0 &          0 &          0 &          1 \\
Electrical Eng.     &      0.044 &     0.206  &          0 &          0 &          0 &          1 \\
Finance             &      0.005 &     0.069  &          0 &          0 &          0 &          1 \\
Mathematics         &      0.198 &     0.398  &          0 &          0 &          0 &          1 \\
Physics             &      0.358 &     0.479  &          0 &          1 &          0 &          1 \\
Statistics          &      0.028 &     0.164  &          0 &          0 &          0 &          1 \\
Female              &      0.096 &     0.193  &          0 &      0.125 &          0 &          1 \\
Male                &      0.481 &     0.388  &          0 &      0.846 &          0 &          1 \\
Num\_papers         &     20.898 &    29.290  &        3.4 &         26 &          0 &      202.5 \\
Yoe                 &      8.543 &     5.346  &      4.667 &       11.5 &          0 &       28.5 \\
Num\_papers\_norm   &     -0.000 &     1.000  &     -0.597 &      0.174 &     -0.713 &      6.199 \\
Yoe\_norm           &     -0.000 &     1.000  &     -0.725 &      0.553 &     -1.598 &      3.732 \\
\hline
Observations        &    627,225 &            &            &            &            &            \\
\hline\hline
\end{tabular}
\end{table}

\end{document}